%% file: paper.tex
\documentclass[]{gtech}
\usepackage{latexml} % Official PDF/LaTeXML conditional support.

\usepackage{amsmath,amssymb}
\usepackage{multirow,bigdelim}
\usepackage{longtable}
\usepackage{tabularray}
\UseTblrLibrary{booktabs}
\usepackage{wrapfig}
\usepackage[most]{tcolorbox}
\usepackage{xcolor}
\usepackage{url}
\usepackage{colortbl}
\usepackage{caption}
\usepackage{pifont}
\usepackage{booktabs}
\usepackage{makecell}
\usepackage{tabulary}
\usepackage{fontawesome5}
\usepackage{bbding}
\usepackage{multicol}
\usepackage{graphicx}
\usepackage{microtype}
\usepackage{enumitem}
\usepackage{tikz}
\usepackage{placeins}
\usepackage{CJKutf8}
\usepackage{flafter}
\usetikzlibrary{arrows.meta,positioning,fit,calc}
\usepackage{pdfrender}

\newcommand{\modelbold}{%
  \pdfrender{TextRenderingMode=FillStroke,LineWidth=0.66pt}%
}

\usepackage{caption}
\usepackage{tabularx}
\usepackage{array}

\makeatletter
\@ifpackageloaded{natbib}{}{\usepackage[authoryear,round,sort]{natbib}}
\AtBeginDocument{%
  \@ifundefined{NAT@longnamesfalse}{}{\NAT@longnamesfalse}%
  \def\NAT@sort{\@ne}%
  \setcitestyle{authoryear,round,semicolon,aysep={,},yysep={,}}%
  \let\cite\citep
  \let\citet\citep
  
}
\makeatother

\ifdefined\DeclareUnicodeCharacter
  \DeclareUnicodeCharacter{FF0C}{,}
\fi

\renewcommand{\title}[1]{%
  \newcommand{\titlelist}{{\huge\fontfamily{optimistic}\fontseries{m}\selectfont #1}}}

\newcommand{\systemname}{{\fontseries{m}\selectfont\texttt{Realtime-Venus}}}

\newcommand{\omnimodel}[1][]{{\fontseries{m}\selectfont\texttt{#1Realtime-Venus-Omni}}}
\newcommand{\audiomodel}[1][]{{\fontseries{m}\selectfont\texttt{#1Realtime-Venus-Audio}}}
\newcommand{\harnessname}[1][]{{\fontseries{m}\selectfont\texttt{#1Realtime-Venus-Harness}}}

\definecolor{prompt}{HTML}{5F84E4}
\definecolor{img}{HTML}{820100}
\definecolor{omniColor}{HTML}{2F6BFF}
\definecolor{audioColor}{HTML}{00A88F}
\definecolor{delegateColor}{HTML}{8A56AC}
\definecolor{ledgerColor}{HTML}{F2A900}

\newlength\savewidth

\newtcolorbox{designbox}[1][]{
  enhanced,
  breakable,
  colback=blue!2,
  colframe=blue!40!black,
  boxrule=0.5pt,
  arc=2pt,
  left=6pt,
  right=6pt,
  top=5pt,
  bottom=5pt,
  #1
}

\providecommand{\RealtimeVenusEvalSetup}{%
  \centering
  \fontencoding{T1}\fontfamily{ptm}\fontsize{9}{11}\selectfont
  \setlength{\tabcolsep}{3pt}%
  \renewcommand{\arraystretch}{1.12}%
  \captionsetup{justification=centering,singlelinecheck=false,skip=6pt}%
  \setlength{\aboverulesep}{0.45ex}%
  \setlength{\belowrulesep}{0.45ex}}
\providecommand{\RealtimeVenusEvalHead}[1]{{\fontsize{8.5}{10}\selectfont\bfseries\shortstack[c]{#1}}}
\providecommand{\RealtimeVenusEvalNum}[2]{#1\phantom{#2}}
\providecommand{\RealtimeVenusEvalNote}[2][\linewidth]{%
  \par\vspace{3pt}%
  \begin{minipage}{#1}
  \fontsize{8}{10}\selectfont\raggedright
  \textit{Notes.} #2
  \end{minipage}}

\title{\textcolor[HTML]{0369FF}{Realtime-Venus}: A full-duplex interaction system with asynchronous delegation}
\author[\*]{%
  {Venus Team, Ant Group}\\
  Tsinghua University
}

\abstract{%
Natural interaction in digital and physical environments requires continuous perception and timely responses. Spoken dialogue relies on acoustic and linguistic cues, while video interaction also requires grounding the conversation in evolving visual context. We present \systemname{}, a proactive full-duplex interaction system with two separately trained 9B models: \omnimodel[\modelbold]{} for audio--visual interaction and \audiomodel[\modelbold]{} for spoken interaction. Each model serves as a complete conversational frontend, integrating continuous perception, conversational control, and native speech generation through a shared causal timeline for user inputs, model outputs, and delegation events.
A dual-loop runtime coordinates live interaction with background reasoning and tool execution. Foreground interaction continues while \harnessname[\modelbold]{} executes tasks asynchronously and returns results for integration into the ongoing dialogue.
Both models follow a common post-training recipe combining offline understanding, proactive full-duplex trajectories, and delegation workflows.
Among the evaluated online models, \omnimodel{} achieves the highest scores on six of eight video benchmarks, including StreamingBench (70.2\%), OVO-Bench (64.7\%), and Daily-Omni (81.3\%). Across eight audio understanding and spoken question answering benchmarks, \audiomodel{} leads the compared models on MMAU (78.0\%), MMAU-Pro (63.2\%), Llama Questions (83.8\%), and Speech CMMLU (67.8\%), while matching the best VoiceBench AlpacaEval score of 4.81. On Full-Duplex-Bench v1.5, \audiomodel{} responds to 75\% of user interruptions and achieves continuation rates of 97\%, 88\%, and 86\% under backchannels, other-directed speech, and background speech, respectively, exceeding Gemini 3.1 Live and GPT-4o on all three continuation metrics.
}
\definecolor{ProjectLabelColor}{HTML}{000000}
\definecolor{ProjectLinkColor}{HTML}{0064E0}

\newcommand{\ProjectResourceLabel}[1]{%
  {\fontfamily{ptm}\fontseries{b}\selectfont
   \textcolor{ProjectLabelColor}{#1}}%
}
\newcommand{\ProjectResourceURL}[2]{%
  \href{#1}{\textcolor{ProjectLinkColor}{\textnormal{#2}}}%
}
\newcommand{\ProjectResourceIcon}[2]{%
  \href{#1}{\raisebox{-0.15em}{\includegraphics[height=1.05em]{#2}}}%
}
\newlength{\ProjectResourceWidth}
\newcommand{\ProjectLinks}{%
  \begingroup
  \normalfont\fontsize{9.5}{12.5}\selectfont
  \setlength{\parindent}{0pt}%
  \setlength{\parskip}{0pt}%
  \renewcommand{\arraystretch}{1.08}%
  \settowidth{\ProjectResourceWidth}{\textnormal{https://github.com/inclusionAI/Realtime-Venus}}%
  \noindent
  \begin{tabular}{@{}l@{\hspace{0.9em}}l@{}}
    \ProjectResourceLabel{Project:} &
    \makebox[\ProjectResourceWidth][s]{%
      \ProjectResourceURL{https://realtime-venus.github.io/}{https://realtime-venus.github.io/}%
      \hfill
      \mbox{%
        \ProjectResourceLabel{Model:}\hspace{0.25em}%
        \ProjectResourceIcon{https://huggingface.co/inclusionAI/Realtime-Venus}{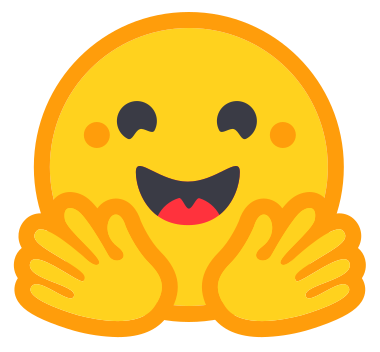}%
        \hspace{0.25em}%
        \ProjectResourceIcon{https://www.modelscope.cn/models/inclusionAI/Realtime-Venus}{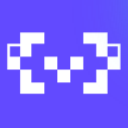}%
      }%
    }\\[4pt]
    \ProjectResourceLabel{Code:} &
    \ProjectResourceURL{https://github.com/inclusionAI/Realtime-Venus}{https://github.com/inclusionAI/Realtime-Venus}%
  \end{tabular}\par
  \endgroup
}

\newlength{\ProjectLogoRaise}
\newcommand{\ProjectFooter}{%
  \noindent
  \begin{minipage}[c]{\dimexpr\linewidth-2.6cm\relax}
    \ProjectLinks
  \end{minipage}%
  \hfill
  \begin{minipage}[c]{2cm}
    \centering
    \raisebox{\ProjectLogoRaise}[\height][\depth]{%
      \includegraphics[width=\linewidth]{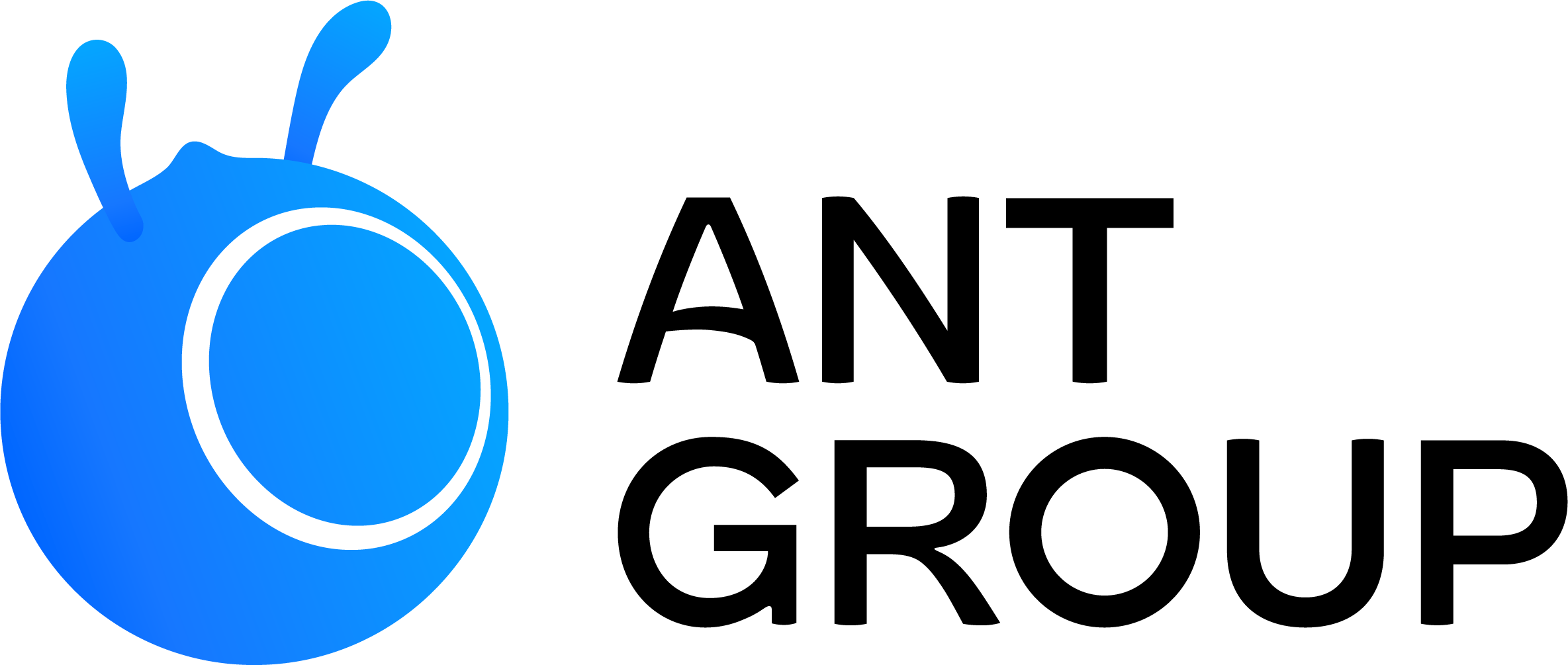}%
    }
  \end{minipage}\par
}

\date{}
\ifdefined\gtechdatalist
  \renewcommand{\gtechdatalist}{}%
\fi
\iflatexml
\else
  \patchcmd{\mymaketitle}
    {\begin{tcolorbox}}
    {\begin{tcolorbox}[overlay={}]}
    {}%
    {\PackageError{realtime-venus}{Could not disable the fixed title logo}%
      {Adapt the title-overlay patch to the current gtech.cls before compiling.}}%
  \renewcommand{\gtechdatalist}{\ProjectFooter}%
\fi

\begin{document}
\maketitle
\iflatexml
  \begin{tcolorbox}[
    colback=cyan!7!white,
    colframe=cyan!7!white,
    boxrule=0pt,
    arc=10pt,
    left=12pt,right=12pt,top=8pt,bottom=8pt
  ]
    \ProjectFooter
  \end{tcolorbox}
\fi

\begin{figure}[!htbp]
    \centering
    \includegraphics[width=\linewidth,height=0.45\textheight,keepaspectratio]{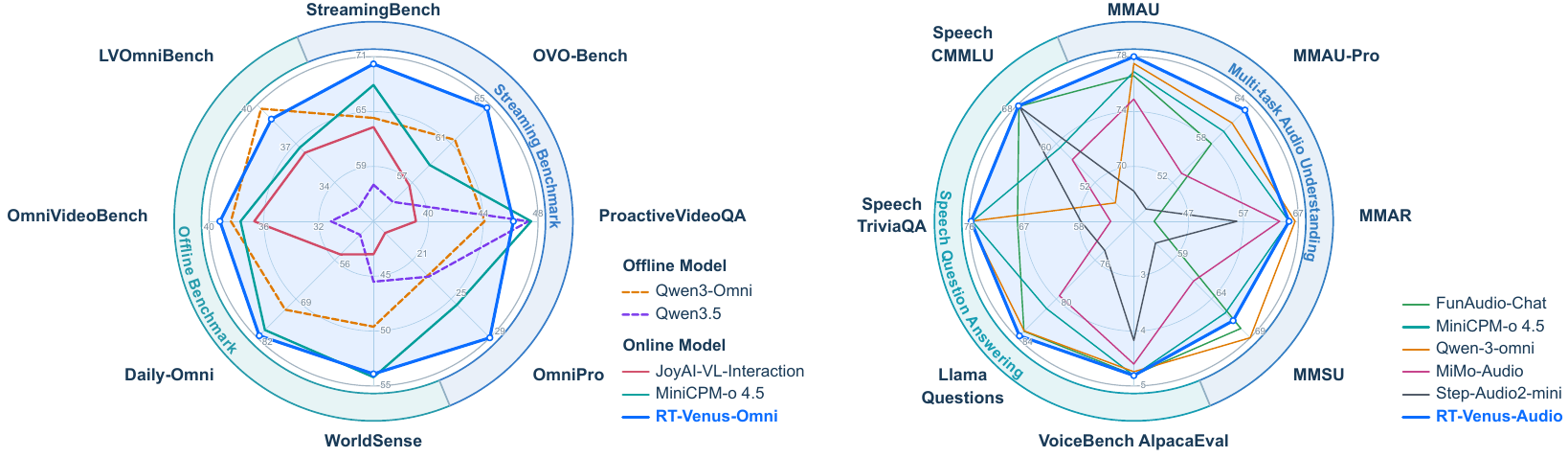}
    \caption{Understanding performance of Realtime-Venus.}
    \label{fig:combined-radars}
\end{figure}

\begin{figure}[!htbp]
    \centering
    \includegraphics[width=\linewidth,height=0.40\textheight,keepaspectratio]{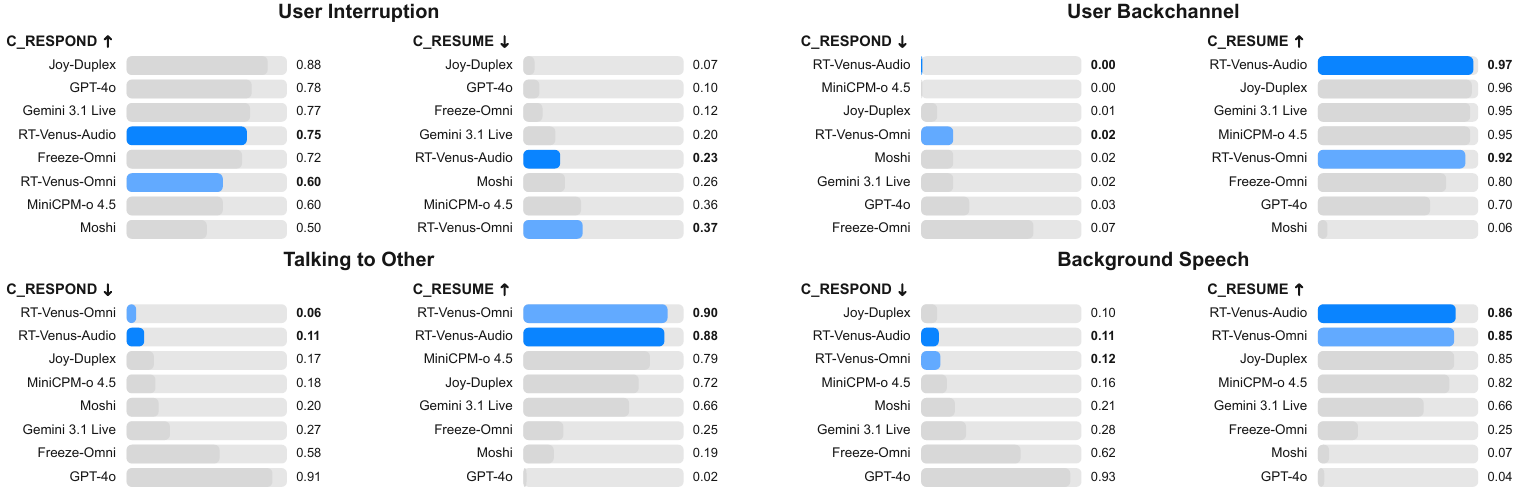}
    \caption{Full-duplex interaction performance of Realtime-Venus.}
    \label{fig:duplex_bench}
\end{figure}
\FloatBarrier
\section{Introduction}
\label{sec:introduction}
Natural interaction requires systems to interpret ongoing observations while deciding when and how to respond, including when to initiate a response without an explicit user request. Recent models increasingly integrate perception, speech generation, and conversational control.
Moshi supports concurrent speech modeling~\citep{defossez2024moshi}; Qwen2.5-Omni and Qwen3-Omni combine multimodal perception with native streaming speech generation~\citep{xu2025qwen2,xu2025qwen3omni}; and MiniCPM-o 4.5 extends these capabilities to proactive full-duplex video interaction~\citep{cui2026minicpmo45realtimefullduplex}. Research on spoken agents explores retrieval, tool calls, and asynchronous external computation during dialogue~\citep{zhang2026duplexsla,chien2026moshirag,huang2026duplexomni,openai2026gptlive}.

Continuous interaction and external computation operate on different timescales within the same session. A background task requires a stable record of the request and its supporting evidence, but its result must be interpreted in a conversation that may have changed during execution. Coordinating task capture with conversational result delivery is essential to maintaining coherent interaction.

We introduce \systemname{}, a proactive full-duplex interaction system that combines native conversational modeling with asynchronous delegation. Built on MiniCPM-o 4.5~\citep{cui2026minicpmo45realtimefullduplex}, the system provides two separately trained models: \omnimodel[\modelbold]{} for audio--visual interaction and \audiomodel[\modelbold]{} for spoken interaction. Each serves as a complete conversational frontend with continuous perception, conversational control, and native speech generation.
A unified streaming formulation aligns user observations, model outputs, private delegation requests, and background results on a shared causal timeline.
Here, the frontend jointly predicts interaction-control tokens, response text, and delegation requests.
The shared policy supports maintaining a response during user backchannels, revising its unspoken continuation after a correction, and initiating background work when a request requires external capabilities.

We pair these models with \harnessname[\modelbold]{}, a shared framework for asynchronous capability execution and result delivery. Within a dual-loop runtime, either frontend maintains live interaction while \harnessname{} manages delegated work in the background.
Each delegation request is bound to its originating session and committed with a snapshot of the evidence available when the request began.
The framework routes the task to a registered capability for asynchronous execution and returns eligible results as private context.
Freshness checks determine result eligibility, while playback-aware delivery governs when results re-enter the conversation.
The frontend then interprets the returned information in the context of the ongoing dialogue and determines the user-facing response.
This design preserves stable context for background execution while allowing the frontend to adapt its response to subsequent changes in user intent.

Training requires trajectories that connect conversational events with interaction decisions and delegated execution. We develop a unified data pipeline combining scenario planning, speech realization, and temporal alignment. Duplex scenarios distinguish backchannels and other-directed speech from interruptions that require stopping, repairing, or redirecting a response. Proactive trajectories supervise when to initiate a response and when to continue listening. Delegation scenarios connect private requests with background execution, returned information, and subsequent responses. The pipeline combines these behaviors within conversations on a common timeline.

The resulting trajectories support a shared post-training recipe that mixes offline understanding, proactive duplex interaction, and delegation workflows. \omnimodel{} uses both audio--visual and audio-only data, whereas \audiomodel{} uses the audio-only subset. Supervision covers interaction-control transitions and subsequent generation, linking decisions about when to listen, speak, or delegate to the response that follows.

The evaluations assess understanding, conversational continuity, and delegation. \omnimodel{} leads the evaluated online models on six of eight video benchmarks, while \audiomodel{} achieves the highest scores in several audio understanding and spoken question answering comparisons. Full-duplex evaluations show high continuation rates under non-interruptive speech. Complementary tool-use and delegation-decision evaluations distinguish correct routing from successful task completion and identify challenges in executing external work during conversation.

Our contributions are summarized as follows:
\begin{itemize}[leftmargin=1.4em,itemsep=2pt,topsep=3pt]
\item \textbf{Proactive full-duplex interaction models.}
We develop \omnimodel[\modelbold]{} and \audiomodel[\modelbold]{}, two separately trained
9B models for proactive audio--visual interaction and full-duplex
spoken dialogue, respectively. Both models integrate continuous
perception, conversational control, native speech generation, and
private delegation under a unified streaming formulation.
To our knowledge, \omnimodel{} is the first full-duplex omni model
to support asynchronous backend invocation for reasoning and tool
execution while maintaining video interaction. With memory augmentation, it also
supports hour-scale video understanding.

    \item \textbf{Asynchronous capability execution with \harnessname[\modelbold]{}.}
    We introduce a shared execution framework that binds tasks to evidence available at the request boundary, executes registered capabilities asynchronously, and returns results to the originating session while preserving frontend control over conversational responses.

    \item \textbf{A coupled duplex and delegation data pipeline.}
    We develop a pipeline combining scenario planning, speech realization, and temporal alignment to construct trajectories coupling conversational events with delegation requests, background results, and response continuations.
\end{itemize}

\section{Related Work}
\label{sec:intro}

\textbf{Omni-modal understanding.}
Omni models integrate text, vision, and audio within shared architectures for understanding and response generation. Gemini~\citep{geminiteam2023gemini} learns from interleaved multimodal data for cross-modal understanding and reasoning, while GPT-4o~\citep{openai2024gpt4o} supports native speech interaction through end-to-end training across text, vision, and audio. Open models extend these capabilities: Baichuan-Omni-1.5~\citep{li2025baichuanomni15} combines multimodal understanding with end-to-end speech generation, and MiniCPM-o 2.6~\citep{openbmb2025minicpmo26} supports continuous audio--visual inputs and streaming speech through online modality processing and time-division multiplexing. Qwen2.5-Omni~\citep{xu2025qwen2} introduces time-aligned audio--visual representations and a Thinker--Talker architecture, while Qwen3-Omni~\citep{xu2025qwen3omni} extends this design with mixture-of-experts components and multi-codebook speech generation. These advances provide modality coverage and streaming output; full-duplex interaction requires incoming observations to influence an ongoing response.

\textbf{Audio understanding and speech generation.}
Audio-language models use complementary approaches to connect acoustic representations with language models. SALMONN~\citep{tang2024salmonn} connects speech and general audio encoders to a large language model (LLM) through a window-level Q-Former. Qwen-Audio~\citep{chu2023qwenaudio} unifies diverse audio understanding tasks through multitask pretraining with hierarchical task tags, while Qwen2-Audio~\citep{chu2024qwen2audio} uses natural-language prompts to support spoken instructions and text-guided audio analysis. SpeechGPT~\citep{zhang2023speechgpt} extends audio-to-text understanding to speech generation by integrating discrete speech representations into an LLM through modality adaptation and cross-modal instruction tuning. More recent models refine this integration: Kimi-Audio~\citep{ding2025kimi} combines continuous acoustic features with discrete semantic tokens, and MiMo-Audio~\citep{xiaomi2025mimoaudio} uses patch-based audio encoding and decoding with next-token pretraining. Fun-Audio-Chat~\citep{tongyi2025funaudiochat} combines dual-resolution speech representations with staged post-training and model merging, while Step-Audio 2~\citep{wu2025stepaudio2} incorporates interleaved text--audio generation, reasoning-oriented reinforcement learning, and external retrieval. Our audio frontend aims to retain acoustic and semantic competence while jointly learning conversational control and delegation.

\textbf{Proactive and full-duplex interaction.}
Full-duplex models address conversational timing by processing incoming speech during response generation. Moshi~\citep{defossez2024moshi} models user and assistant audio as parallel streams and couples text with speech through its Inner Monologue formulation. Freeze-Omni~\citep{wang2025freezeomni} uses an auxiliary classifier on LLM hidden states, trained with a joint state-prediction and language-modeling objective, to predict chunk-level dialogue states and determine whether incoming speech requires interrupting the ongoing response. Fun-Audio-Chat~\citep{tongyi2025funaudiochat} extends joint speech--text modeling with parallel input streams and synthesized concurrent dialogues. JoyAI-Talker~\citep{bai2026joyaitalker} combines a modular Thinker--Talker architecture with Joy-Duplex for interaction-state prediction and turn control. In multimodal settings, MiniCPM-o 4.5~\citep{cui2026minicpmo45realtimefullduplex} introduces Omni-Flow to align incoming audio--visual streams with outputs for simultaneous perception, speech, and proactive engagement. Video-focused methods study when observations warrant a response: LiveStar~\citep{yang2025livestar} combines response--silence decoding with memory-aware streaming inference, while MMDuet2~\citep{wang2025mmduet2} learns response timing and content through multi-turn reinforcement learning. Our formulation extends native streaming interaction by placing conversational control, response generation, and private delegation within a shared policy and timeline.

\textbf{Tool use and asynchronous delegation.}
Tool-augmented language models connect reasoning with external capabilities. ReAct~\citep{yao2023react} interleaves reasoning and actions with environmental feedback. Spoken and multimodal systems extend this approach to ongoing interaction. DuplexSLA~\citep{zhang2026duplexsla} jointly models speech and a structured action channel for planning, interaction control, and tool calls. MoshiRAG~\citep{chien2026moshirag} provides selective asynchronous retrieval for a full-duplex speech model, while DuplexOmni~\citep{huang2026duplexomni} pairs continuous multimodal interaction with a pluggable asynchronous thinking layer. JoyAI-VL-Interaction~\citep{joyai2026vlinteraction} connects proactive visual interaction to background delegation using external automatic speech recognition (ASR) and text-to-speech (TTS) components. GPT-Realtime~\citep{openai2025gptrealtime} supports asynchronous function calling, and NemotronLabs VoiceChat~\citep{nvidia2026voicechat} provides a separate output channel for tool-calling scripts. We study a private delegation interface for separately trained audio and omni frontends. \harnessname{} clarifies execution boundaries by fixing evidence at the request boundary, executing registered capabilities, and returning results to the originating session for integration into the current dialogue.

\section{System Overview}
\label{sec:system-overview}

\begin{figure}[!htbp]
    \centering
    \includegraphics[
        width=\linewidth,
        height=0.50\textheight,
        keepaspectratio
    ]{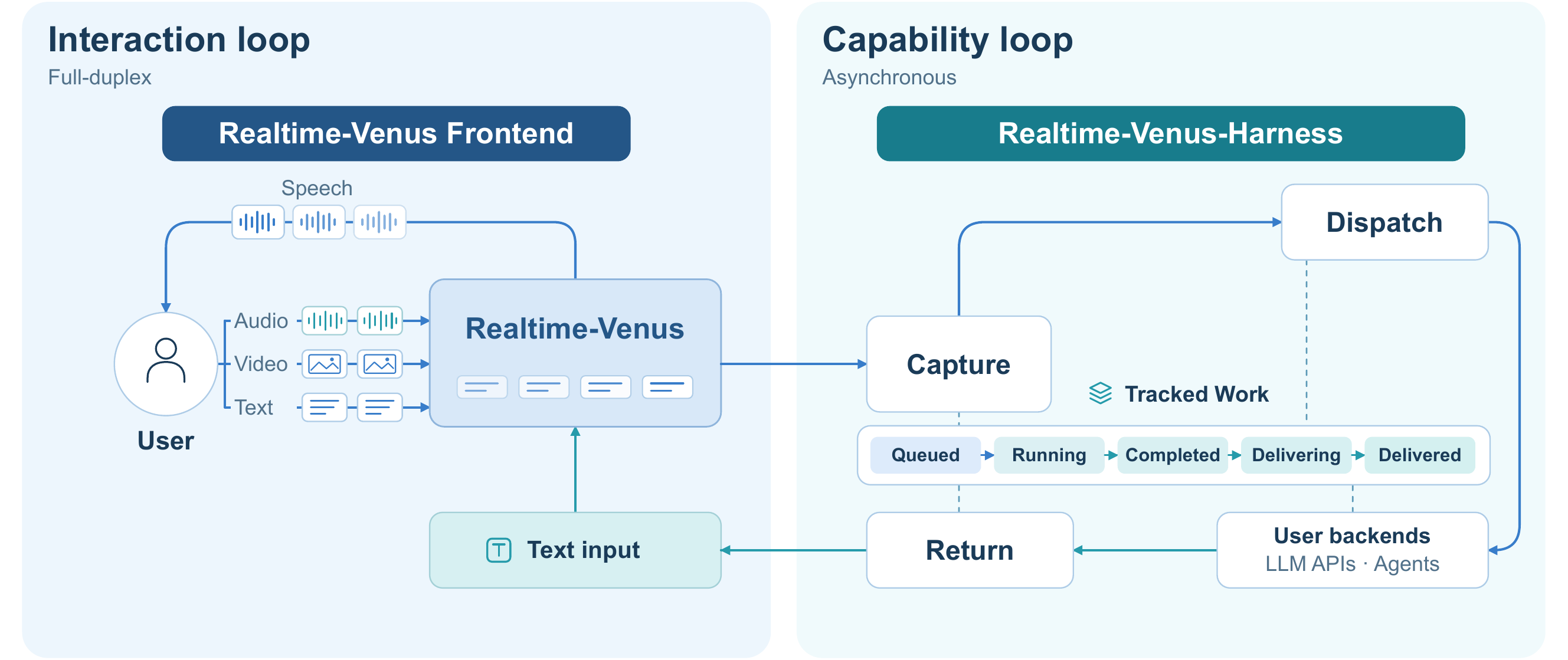}
    \caption{\systemname{} runtime: full-duplex interaction with asynchronous delegation.}
    \label{fig:Realtime-Venus-overview}
\end{figure}

\systemname{} combines a full-duplex conversational frontend with \harnessname{} for asynchronous capability execution and reply preparation. Each session uses either \audiomodel{}, which processes continuous audio, or \omnimodel{}, which additionally processes visual inputs. Both share the same delegation interface: the frontend handles live interaction and speech output, while the harness executes delegated tasks and prepares replies. Figure~\ref{fig:Realtime-Venus-overview} summarizes the architecture.

\subsection{Dual-loop architecture}
\label{sec:dual-loop-architecture}

\textbf{Interaction loop.} The frontend continuously processes incoming media, updates the session state, and controls when to listen, speak, or yield. It handles requests that can be answered directly and provides streaming text and speech output through its native Thinker--Talker architecture. Perception remains active during speech and background task execution.

\textbf{Capability loop.} A private natural-language delegation request from the frontend activates this loop. The host hides the request span from user-facing output, captures the context available at the request boundary, and creates an asynchronous work item. The harness selects a backend capability, executes the task, and polishes the result into a reply for spoken delivery. Eligible replies return to the originating session through the private background channel. The frontend chooses when to speak the prepared text during the ongoing interaction; the harness determines its content and wording.

\begin{figure}[!h]
    \centering
    \includegraphics[
        width=\linewidth,
        height=0.58\textheight,
        keepaspectratio
    ]{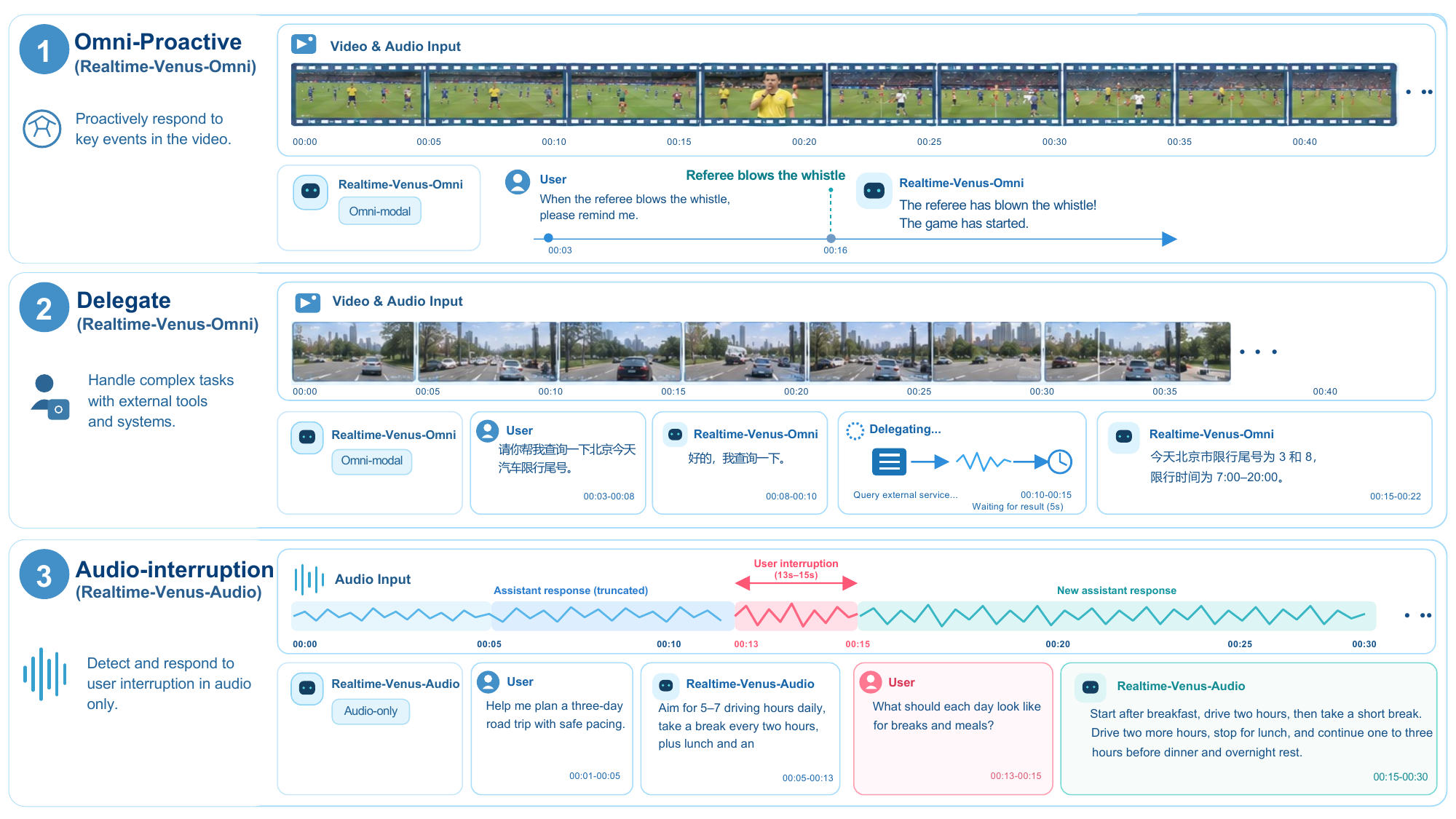}
    \caption{Example interactions with \systemname{}.}
    \label{fig:Realtime-Venus-case}
\end{figure}

The interaction loop runs continuously and is latency-sensitive, while the capability loop handles tasks on demand over potentially longer durations. Their shared interface exchanges a task package bounded by the request's causal context and a prepared reply bound to the originating session, allowing asynchronous execution alongside live interaction, as illustrated in Figure~\ref{fig:Realtime-Venus-case}.

\subsection{Unified runtime abstraction}
\label{sec:unified-runtime-abstraction}

Both frontends follow the same runtime structure: they receive media and newly returned replies, update retained session state, and determine interaction behavior and output. Consider a session $\sigma$ with frontend $m\in\{\mathrm{audio},\mathrm{omni}\}$, and let $k$ index one-second chunks; session superscripts are omitted. The media inputs are $x_k^{\mathrm{audio}}=u_k$ and $x_k^{\mathrm{omni}}=(u_k,v_k)$, where $u_k$ contains causal audio features and $v_k$ contains aligned visual features, with $v_k=\varnothing$ when no frame is available.

Let $s_k$ denote the frontend state retained before chunk $k$, including model context, partial output spans, and previously admitted reply text awaiting delivery. The background input $b_k$ contains private reply text newly admitted before assistant generation in that chunk. Previously received replies remain available through the retained state, so their delivery does not require a new background message in every chunk. The runtime feedback $\eta_k$ records causal events such as playback acknowledgments. The output $O_k=(c_k,Y_k,D_k,S_k)$ comprises interaction-control tokens, foreground text, delegation text, and speech tokens; only $Y_k$ and $S_k$ are user-facing. The causal runtime transition is
\begin{equation}
    (s_{k+1},O_k)
    =F_m(s_k,x_k^m,b_k;\eta_k).
    \label{eq:frontend-runtime}
\end{equation}
Here, $F_m$ summarizes streaming decoding and runtime scheduling while preserving the causal input--output order within each chunk, as specified in Section~\ref{sec:stream-serialization}. Delegation text $D_k$ may extend across multiple chunks. A completed request commits a work item whose evidence boundary was fixed when the request began, and its prepared reply can be admitted at a later input boundary. For delegated replies, the frontend uses the current interaction state to schedule speech output of the text prepared by the harness. Section~\ref{sec:delegate-design} details task capture, capability execution, and reply delivery.

\FloatBarrier

\begin{figure}[t]
    \centering
    \includegraphics[width=1\linewidth]{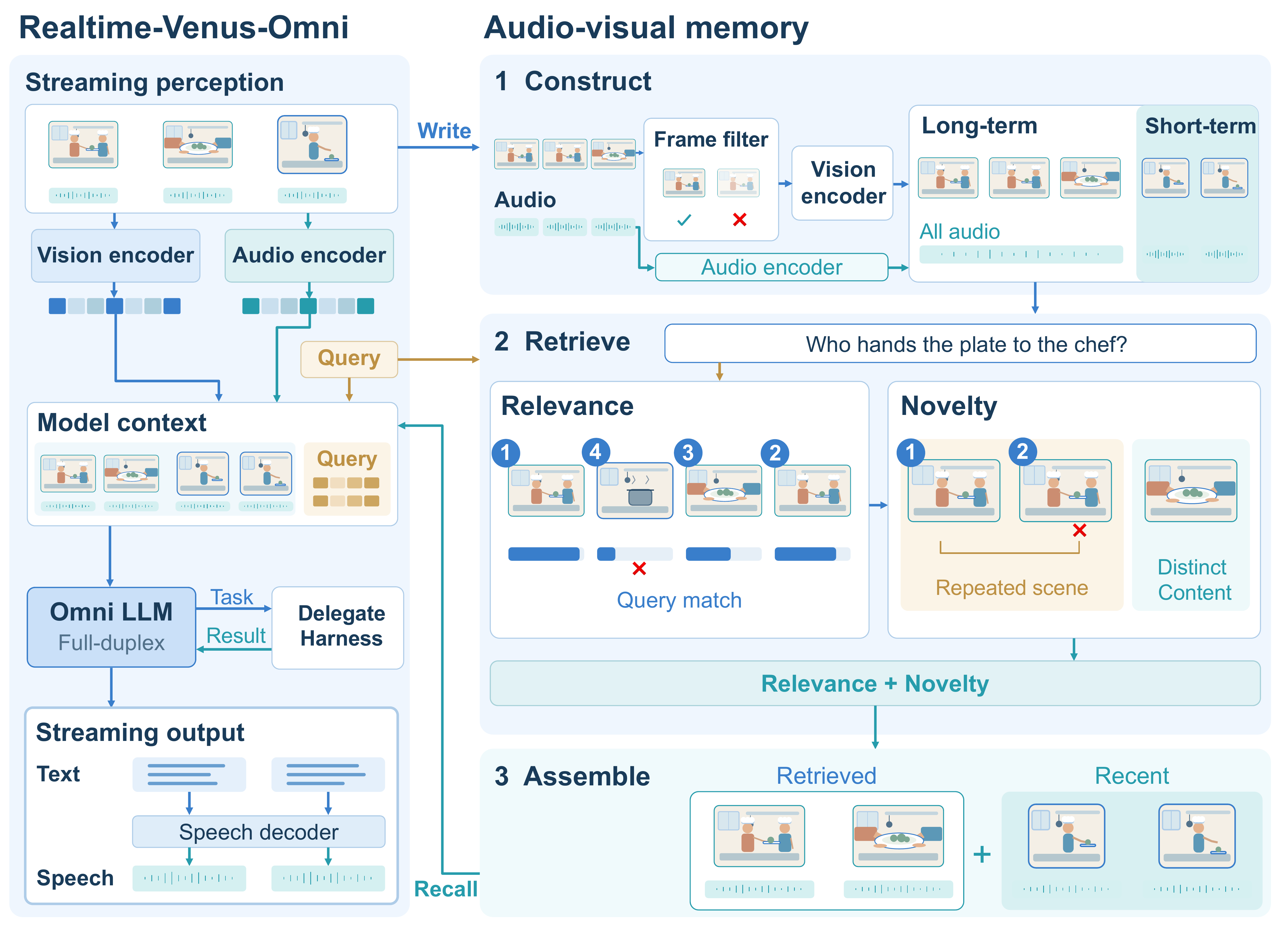}
    \caption{Architecture of \omnimodel{} for audio-visual perception, full-duplex interaction, and asynchronous delegation.}
    \label{fig:intrust-omni}
\end{figure}

\section{Model Design} \label{sec:omni}
\input{sections_omni/omni_model}

\subsection{Unified stream serialization}
\label{sec:stream-serialization}

\begin{figure}[!htbp]
    \centering
    \includegraphics[width=\linewidth,height=0.48\textheight,keepaspectratio]{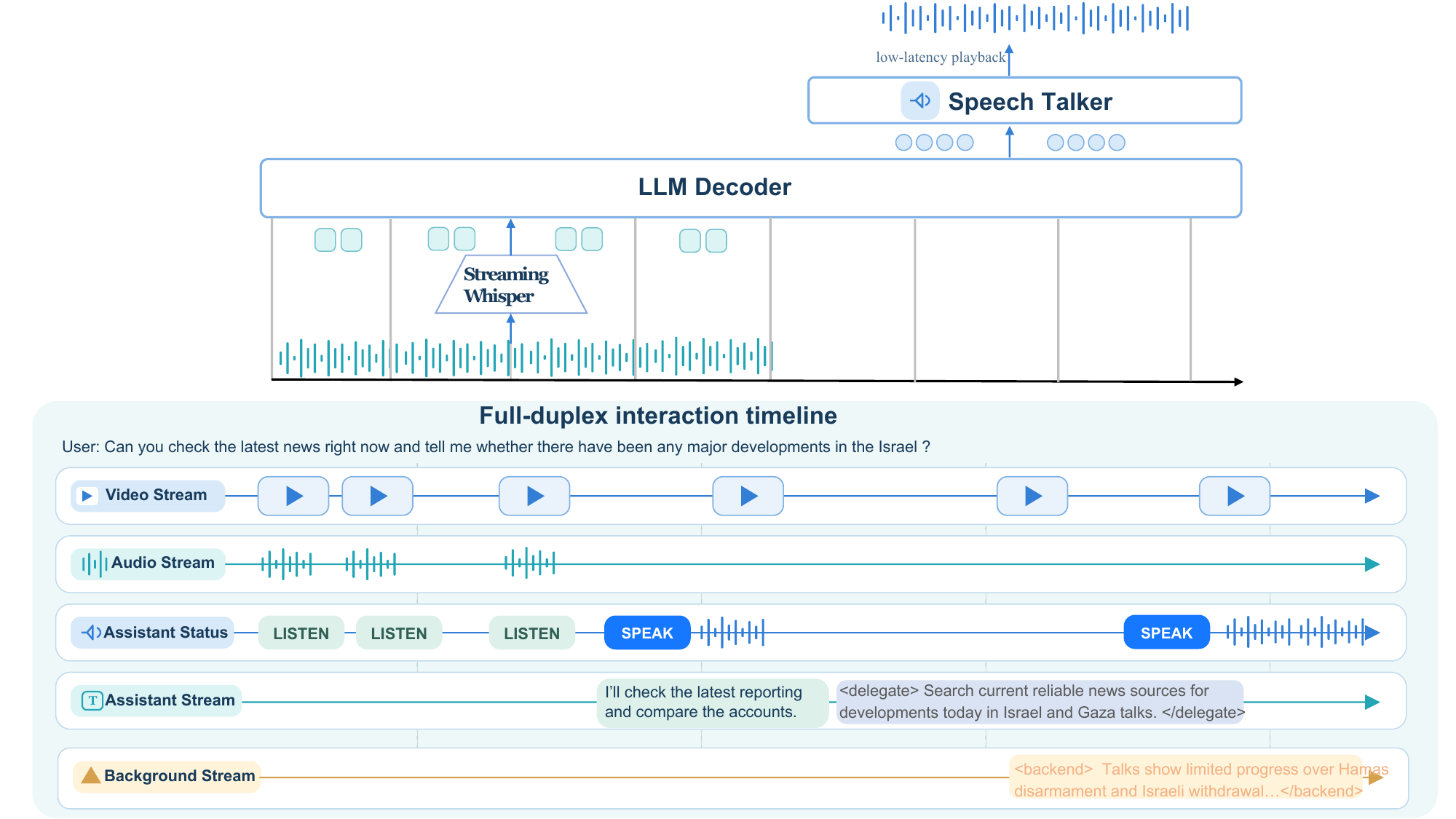}
    \caption{Unified stream serialization for concurrent perception, speech generation, and delegation.}
    \label{fig:duplex-stream}
\end{figure}

\omnimodel{} and \audiomodel{} represent streaming interaction as a sequence of one-second chunks, with temporal boundary $t_k=k\,\mathrm{s}$ and chunk $k$ covering $[t_{k-1},t_k)$. Each chunk aligns three logical streams---the user stream, assistant stream, and background stream---on a shared conversational clock, as illustrated in Figure~\ref{fig:duplex-stream}.

\textbf{User stream.} The user stream is a causal sequence of time-aligned perceptual features. In \omnimodel{}, projected features from ViT~\citep{dosovitskiy2020image} are interleaved with projected features from Whisper~\citep{Whisper} at approximately 10 audio features per second; \audiomodel{} omits the visual features. The stream remains active during assistant output, preserving visual events, pauses, feedback, and overlapping speech on the shared timeline.

\textbf{Assistant stream.} The assistant stream combines foreground text, aligned S3 speech tokens~\citep{du2024cosyvoice,du2024cosyvoice2} at about 25 tokens per second, and optional text-only delegation instructions. LLM hidden states preserve contextual prosody during speech generation, while delegation instructions are routed to the backend rather than spoken.

\textbf{Background stream.} The background stream is an asynchronous text-only sequence that connects the interaction models to a more capable backend agent for complex reasoning and tool-based tasks. A delegation instruction launches backend computation, which incurs a variable response delay. A pending task contributes no result tokens, allowing foreground interaction to continue. Once eligible for delivery, its result is inserted before the assistant stream at an available chunk boundary as external context rather than a prediction target.

The three streams are interleaved into a single token stream consumed by the LLM backbone. The serialization takes the following form:
\begin{tcolorbox}[
    colback = gray!6,
    colframe = black!60,
    boxrule = 0.6pt,
    rounded corners,
    left = 4pt, right = 4pt,
    top = 2pt, bottom = 2pt,
    fontupper = {\ttfamily\small},
    halign = left
]
\newcommand{\ChunkInput}[1]{{%
    \fontencoding{T1}\fontfamily{ptm}\fontseries{m}\fontshape{it}\selectfont
    #1%
}}

chunk 5: <unit> \ChunkInput{V A}
$\rightarrow$ \textbf{<|listen|>} </unit>\\[2pt]

chunk 6: <unit> \ChunkInput{V A}
$\rightarrow$ \textbf{<|speak|> T <|chunk\_eos|><|turn\_eos|>} </unit>\\[2pt]

chunk 7: <unit> \ChunkInput{V A}
$\rightarrow$ \textbf{<|speak|> <delegate> T </delegate><|chunk\_eos|><|turn\_eos|>} </unit>\\[2pt]

chunk 8: <unit> \ChunkInput{V A}
$\rightarrow$ \textbf{<|listen|>} </unit>\\[2pt]

...\\[2pt]

chunk 13: <unit> \ChunkInput{V A}
$\rightarrow$ \textbf{<|listen|>} </unit>\\[2pt]

chunk 14: <unit> \ChunkInput{V A <backend> T </backend>}
$\rightarrow$ \textbf{<|speak|> T <|chunk\_eos|><|turn\_eos|>} </unit>

\end{tcolorbox}
\RealtimeVenusEvalNote{Each line represents one unit, written as \texttt{<unit>} \textit{input} $\rightarrow$ \textbf{output} \texttt{</unit>}. Outputs are shown within one chunk for clarity; in practice, the text output tokens (\textbf{T}) of a single turn may span multiple consecutive chunks.}

Here, $T$ denotes text tokens, $A$ an audio representation, and $V$ a projected visual representation. The $V$ and $A$ tokens are interleaved according to their arrival times in \omnimodel{}, whereas \audiomodel{} serializes only $A$. 

External tool requests appear as delegate spans, and their results appear as background spans on the shared timeline. We add four control tokens---\mbox{\normalfont\detokenize{<delegate>}}, \mbox{\normalfont\detokenize{</delegate>}}, \mbox{\normalfont\detokenize{<backend>}}, and \mbox{\normalfont\detokenize{</backend>}}---to the tokenizer, input embeddings, and LM head. When tools are required, the model emits a self-contained natural-language task inside a delegate span, which is hidden from display and speech. The backend agent decomposes the task into one or more function calls and returns eligible results through a background span at an available input boundary. An additional request arriving during assistant speech can therefore be dispatched without closing the foreground turn, while compound requests can trigger multiple calls in semantic order. In both cases, tool execution proceeds asynchronously alongside foreground speech and perception.

Three state tokens represent full-duplex turn control. \mbox{\normalfont\detokenize{<|listen|>}} indicates continued perception without speech, \mbox{\normalfont\detokenize{<|speak|>}} marks active foreground generation, and \mbox{\normalfont\detokenize{<|turn_eos|>}} closes a completed assistant turn. Together, they provide a compact chunk-level interface for interpreting backchannels, distinguishing interruptions, and coordinating transitions between listening and speaking.

To synchronize generated speech with the conversational clock, we adapt text emission to accumulated playback progress. Let $\tau_{k-1}$ denote the end time of previously scheduled speech and $\mathcal{D}(Y_{k,1:n})$ the estimated duration of the first $n$ candidate text tokens. At chunk $k$, we select
\begin{equation}
    n_k=\arg\min_n\left|\tau_{k-1}+\mathcal{D}(Y_{k,1:n})-t_k\right|,
    \label{eq:stream-alignment}
\end{equation}
where $t_k$ is the current chunk boundary and $\tau_{k-1}$ carries the accumulated playback progress. The scheduler emits fewer text tokens when speech lags and more when playback capacity is available, aligning the response with the latest user audio and background results. During training, text tokens and their corresponding S3 tokens are assigned to chunks by start time, teaching history-dependent interleaving without a fixed text-speech ratio.

\FloatBarrier
\subsection{Full-duplex interaction}
\label{sec:native-full-duplex}

We formulate spoken interaction as chunk-level autoregressive modeling over the three synchronized streams. Let $U=(u_1,\ldots,u_K)$ denote the causal user-audio features, $B=(b_1,\ldots,b_K)$ the sparse background results available at each chunk, and $O_k=(c_k,Y_k,D_k,S_k)$ the assistant output at chunk $k$. Here, $c_k$ is an interaction-control token, $Y_k$ is foreground response text, $D_k$ is an optional delegate request, and $S_k$ is the aligned S3 speech-token segment. Fields that are inactive in a chunk are represented as empty sequences.

The LLM with parameters $\theta$ predicts interaction control, foreground text, and delegate text, while the speech decoder with parameters $\phi$ generates speech tokens from the LLM hidden states $H$. The joint policy factorizes as
\begin{equation}
    p_{\theta,\phi}(O_{1:K}\mid U,B)
    =\prod_{k=1}^{K}
      p_{\theta}\!\left(c_k,Y_k,D_k\mid U_{\leq k},B_{\leq k},O_{<k}\right)
      p_{\phi}\!\left(S_k\mid H_{\leq k},Y_{\leq k},S_{<k}\right).
    \label{eq:native-duplex-policy}
\end{equation}
This factorization keeps acoustic generation outside the main LLM while allowing speech, turn control, and delegation to share the same semantic representation. Background results are external observations rather than model outputs and become available only after the corresponding backend computation finishes.

The control variable $c_k$ determines whether the assistant listens, speaks, or closes its current turn. Because $c_k$ is predicted from the same semantic representation used for response generation, overlapping audio is interpreted according to its conversational role rather than treated as a uniform stop signal. We distinguish the following cases:
\begin{itemize}
    \item \textbf{Pauses and background noise.} Silence, hesitation, or acoustic activity not directed at the assistant produces \mbox{\normalfont\detokenize{<|listen|>}}, keeping perception continuously active without prematurely starting or terminating a response.
    \item \textbf{Backchannels.} A short acknowledgment such as ``yes'' or ``right'' does not claim the conversational floor. The model preserves \mbox{\normalfont\detokenize{<|speak|>}} and continues the current response plan.
    \item \textbf{Interruptions.} When overlapping user speech takes the conversational floor with a correction, redirection, or new request, the model emits \mbox{\normalfont\detokenize{<|turn_eos|>}} to terminate the stale response. It then incorporates the new intent into its conversational state before generating the next \mbox{\normalfont\detokenize{<|speak|>}} segment or delegate request, revising the unspoken continuation based on the new input while leaving already played audio unchanged.
\end{itemize}

We model interruptions as event-conditioned semantic control and jointly supervise the state transition and subsequent generation trajectory. The \emph{stop/reject} action closes the turn and discards its unspoken remainder; \emph{repair/update} incorporates new evidence and regenerates the stale continuation; and \emph{redirect/follow-up} suspends the current plan and transfers control to a new request. A shared semantic state across turn control, text, speech, and delegate generation enables acoustically similar overlaps to produce distinct continuations, supporting native full-duplex control beyond binary reactions triggered by voice activity detection (VAD).

\subsection{Training data} \label{ssec:omni-data}

\omnimodel{} and \audiomodel{} are trained on a common post-training corpus of over 2.8 million samples covering nine data categories, summarized in Table~\ref{tab:omni-data}. The corpus is organized into video and audio data. \omnimodel{} uses both modalities, while \audiomodel{} is trained only on audio data. The data cover three main capabilities: \emph{offline understanding}, including general audio--visual understanding, general audio understanding, and spoken question answering; \emph{proactive duplex interaction}, including visual-, multimodal-, speech-, and audio-driven interaction with response timing and interruption handling; and \emph{delegation}, which trains the delegate--backend--restate workflow described in Section~\ref{sec:delegate-design}.

\begin{table}[htbp]
\centering
\small
\caption{Training data composition.}
\label{tab:omni-data}
\setlength{\tabcolsep}{4pt}
\begin{tabularx}{\linewidth}{
    @{}
    >{\raggedright\arraybackslash}p{0.13\linewidth}
    >{\raggedright\arraybackslash}p{0.29\linewidth}
    >{\centering\arraybackslash}p{0.10\linewidth}
    >{\raggedright\arraybackslash}X
    @{}
}
\toprule
\textbf{Modality} & \textbf{Category} & \textbf{Size} & \textbf{Role} \\
\midrule
\multirow{5}{*}{Video}
 & General AV understanding & 936k & offline AV comprehension \\
 & Visual-proactive duplex & 454k & visually triggered proactivity \\
 & Omni-proactive duplex & 201k & multimodal proactive interaction \\
 & Speech-in duplex & 315k & proactive response to in-stream spoken queries \\
 & Delegate & 95k & delegate--backend--restate \\
\midrule
\multirow{4}{*}{Audio}
 & General audio understanding & 470k & offline audio comprehension \\
 & Spoken question answering & 205k & offline spoken question answering \\
 & Speech-only duplex & 100k & full-duplex interaction and interruption handling \\
 & Delegate & 90k & audio delegation \\
\bottomrule
\end{tabularx}
\end{table}

By modality, the video group constitutes approximately 70\% of the
corpus and is consumed exclusively by \omnimodel{}, whereas the audio
group accounts for approximately 30\% and is shared by \omnimodel{} and 
\audiomodel{}. Offline understanding represents approximately
56\% of the corpus, proactive duplex interaction accounts for approximately
37\%, and delegation workflows comprise the remaining 6\%. Both duplex and
delegation sources contain sessions with interruptions, so overlapping speech is
supervised as part of ordinary interaction rather than through a separate module.
Section~\ref{sec:data} describes how the full-duplex and delegate datasets are constructed. Training uses this mixture under the unified recipe
described in Section~\ref{ssec:omni-training}.

\subsection{Training recipe} \label{ssec:omni-training}

\omnimodel{} and \audiomodel{} use a unified post-training recipe
for multimodal streaming. The recipe supports video, audio-only, and text-only
inputs and combines full-duplex and offline turn-based formats in a single
post-training pass. It mixes proactive duplex data (including speech-in queries),
delegation data, and general understanding data, as detailed in
Section~\ref{ssec:omni-data}. The two variants share the training recipe but
differ in input modalities and training-data coverage. The user stream of
\omnimodel{} interleaves projected visual and audio tokens, and the model uses
both video and audio data. In contrast, \audiomodel{} retains only the audio
segments and is trained on the audio group.

\textbf{Objective.} Supervision is sparse: loss is computed only on response
spans, excluding system, user, and media-placeholder tokens. In offline
multi-turn data, assistant history is supervised together with the final
turn. For full-duplex data, supervision covers the per-second
\texttt{<|listen|>}/\texttt{<|speak|>} decision tokens together with the spoken
text, and a long reply is supervised continuously across multiple units.
Loss is normalized per sample: each sample's weighted loss is divided by its
total supervision weight before averaging over the batch, preventing its
contribution from scaling with the number of supervised tokens. For long answers, the LM-head
cross-entropy is evaluated in chunks with recomputation, bounding activation
memory without changing the objective. Samples that fail to decode or exceed the
length budget are discarded during preprocessing and excluded from training.
Only the Thinker is updated. The acoustic decoder, which synthesizes audible
speech from Thinker hidden states, remains fixed and is excluded from the
training objective.

\FloatBarrier
\section{Harness Design}
\label{sec:delegate-design}

\harnessname{} connects full-duplex interaction to models and tools for asynchronous reasoning and execution. It provides \omnimodel{} and \audiomodel{} with a common execution interface, manages each task's context, and tracks its progress through delivery to the conversation. Background execution uses evidence fixed at the request boundary, while the frontend interprets the returned result using the conversational state at reintegration.

\begin{figure}[!htbp]
    \centering
    \includegraphics[width=\linewidth,height=0.52\textheight,keepaspectratio]{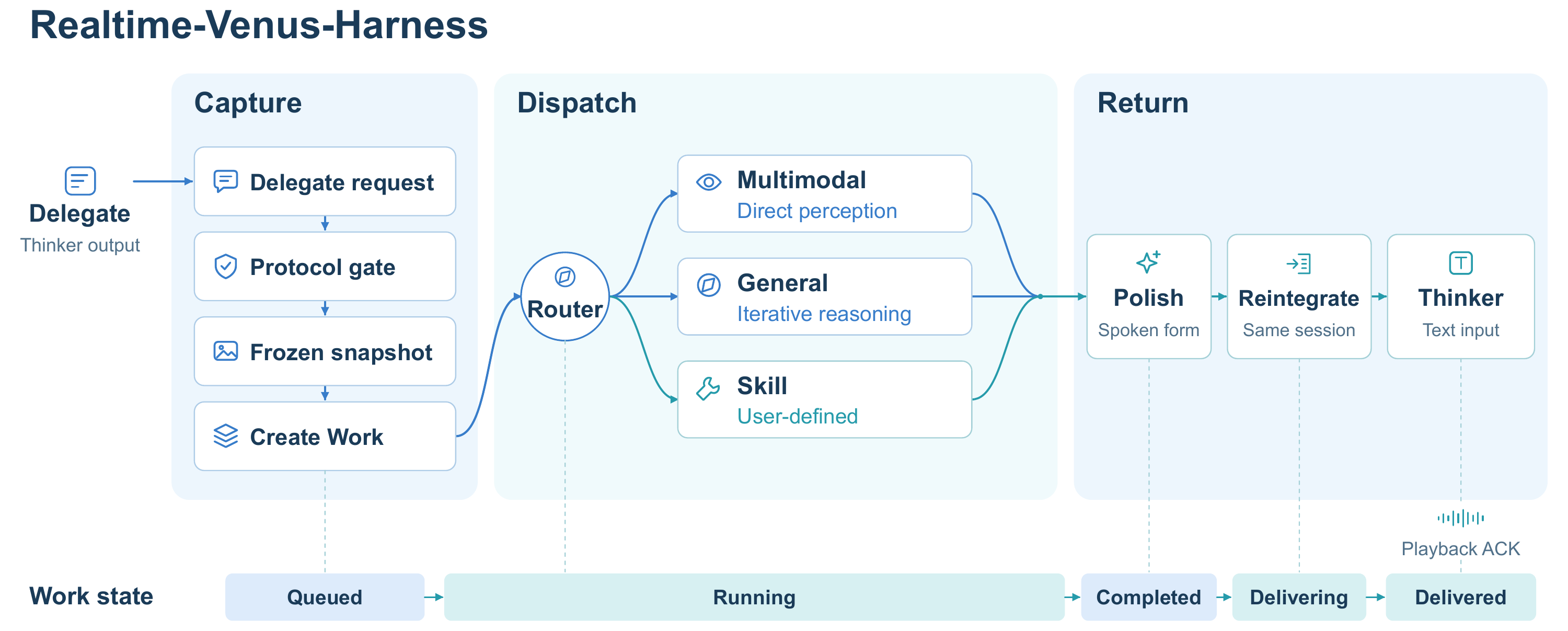}
    \caption{Architecture of \harnessname{} for asynchronous task execution and result delivery.}
    \label{fig:delegate-harness}
\end{figure}

As shown in Figure~\ref{fig:delegate-harness}, \harnessname{} comprises three stages: \emph{capture}, \emph{dispatch}, and \emph{return}, with work lifecycle tracking across all three. Capture establishes the task and its evidence scope; dispatch selects and invokes a registered capability; return prepares the result for the originating frontend. A work item preserves task identity and records progress throughout execution and delivery, while foreground perception and interaction continue independently.

\subsection{Causal task capture}
\label{sec:causal-work}

The frontend specifies an objective through the private span \texttt{<delegate>}~$q_i$~\texttt{</delegate>}. The host extracts this natural-language request from the raw Thinker token stream and suppresses it from visible text and speech. Protocol validation and task-identity checks ensure that an accepted request creates at most one work item. The frontend can therefore request external help without encoding backend-specific tool calls or capability schemas.

Let $i$ index accepted request identities in session $\sigma$, and let $\kappa_i=(T_i,n_i)$ contain the opening token's timestamp and the latest input sequence observed at that boundary. Capture uses the bounded session buffer $\mathcal{B}^{\sigma}$ to construct
\begin{equation}
    \begin{aligned}
    X_i &= \operatorname{Snap}(\mathcal{B}^{\sigma};\kappa_i,\Delta),\\
    w_i &= P(i,\sigma,q_i,X_i),
    \end{aligned}
    \label{eq:harness-runtime}
\end{equation}
where $\Delta$ is the look-back horizon and $P$ is a stateful registration operation keyed by request identity. Registration occurs only after the closing token completes $q_i$. Snapshot timestamps and $\Delta$ use a common time base; timestamp and sequence fences restrict evidence to observations available at $T_i$, with media clipped to the permitted window. Prior assistant speech requires playback confirmation by $T_i$. Later observations or acknowledgments cannot alter $X_i$. The work item retains this evidence boundary independently of its commitment time.

\subsection{Extensible capability execution}
\label{sec:capability-adaptive-execution}

After capture, dispatch assigns each work item to a registered backend capability. A shared execution interface allows the frontend to express a natural-language objective while the harness manages capability selection and execution.

The registry provides two core capabilities and supports extensions through registered skills. The \emph{multimodal} capability uses frozen audio or visual evidence for perception and question answering. The \emph{general} capability returns structured plans for open-ended, multi-step objectives. A registered \emph{skill} invokes a domain executor for task-specific procedures and tools. Each capability has a contract specifying its permitted inputs and argument schema. 

Let $\mathcal{C}=\{\gamma_{\mathrm{mm}},\gamma_{\mathrm{gen}}\}\cup\mathcal{C}_{\mathrm{skill}}$ denote the registry and $\mathcal{A}_{\gamma}$ the valid argument space of capability $\gamma$. Given an objective $q_i$, the router $R$ makes one model-based decision using the registered contracts and $z_i$, which contains a compact evidence summary and permitted request metadata. It selects a capability and supplies its arguments without inspecting raw media. A validated decision satisfies
\begin{equation}
    (\gamma_i,\alpha_i)=R(q_i,z_i,\mathcal{C}),
    \qquad \gamma_i\in\mathcal{C},
    \quad \alpha_i\in\mathcal{A}_{\gamma_i}.
    \label{eq:harness-routing}
\end{equation}
Schema validation rejects invalid selections or arguments before execution.

On successful execution, the selected executor produces a normalized result $r_i=E_{\gamma_i}(w_i,\alpha_i)$. This common result format allows all execution paths to feed the same return stage. New skills extend the registry by supplying a capability contract and an executor that follows this interface, preserving the frontend's delegation protocol and the shared result format.
\subsection{Conversational result reintegration}
\label{sec:stateful-result-reentry}

The return stage of \harnessname{} prepares the text to be spoken in response to a delegated task. The adapter $A$ applies a shared text-only polishing pass to turn the execution result into a reply, removes reserved control tokens, and wraps the prepared text in a single private \texttt{<backend>} message:
\begin{equation}
    \widetilde b_i=A(w_i,r_i),
    \label{eq:harness-reintegration}
\end{equation}
where $w_i$ is the originating work item and $r_i$ is its normalized execution result. The harness determines the content and wording of the reply before handing it to the frontend.

The prepared message returns to the originating session through the private background channel. Messages admitted at chunk $k$ form the background input $b_k$ in Eq.~\ref{eq:frontend-runtime}. Admission makes the reply available to the frontend as private conversational context, while spoken delivery may occur later. The frontend uses the ongoing interaction to select a suitable time to speak the prepared text and produces speech through the native Talker. This division assigns reply preparation to the harness and conversational timing and speech output to the frontend.

\subsection{Work lifecycle and delivery tracking}
\label{sec:work-lifecycle}

Successful execution and delivery follow the path
\[
    \textsc{Queued}\rightarrow\textsc{Running}\rightarrow
    \textsc{Completed}\rightarrow\textsc{Delivering}\rightarrow
    \textsc{Delivered}.
\]
A committed work item initially enters \textsc{Queued}; routing, capability execution, and reply polishing occur during \textsc{Running}. The work item enters \textsc{Completed} when the harness has constructed its terminal result, including the prepared reply text. An unrecoverable execution error produces the terminal state \textsc{Failed}. The request carries a freshness deadline; a result marked stale at terminal-result construction is excluded from reintegration.

\textsc{Completed} records that the reply has been prepared, while \textsc{Delivering} begins when the prepared message is reserved for admission to the originating session. The frontend controls when the reply is spoken. For a speech-bearing reply, \textsc{Delivered} requires both completed frontend generation for that reply and playback acknowledgment of every associated Talker chunk. These states distinguish a reply that is ready for delivery from one whose speech has been played.

This lifecycle tracks each work item through execution and delivery, providing a basis for handling exceptions at different stages. When an exception occurs, the recorded state helps the runtime identify the affected stage and determine how to handle the task.

\FloatBarrier

\section{Full-duplex and Delegation Data Pipeline}
\label{sec:data}
Full-duplex interaction involves concurrent user, assistant, and environmental streams. Acoustically similar events may require different responses depending on their conversational function: a brief backchannel
should normally preserve the current response, whereas an interruption may require the assistant to stop and revise it. We therefore represent each session on a shared timeline that records speaker activity,
overlap, intended addressee, conversational function, and the state of the active assistant response. The same representation connects interaction events to delegation operations, enabling conversational control and
background computation to be supervised within a unified trajectory. This continuous-stream formulation follows prior work on real-time spoken
dialogue~\citep{defossez2024moshi}.

\subsection{Unified data construction}
\label{sec:unified-data-construction}

\begin{figure}[!htbp]
    \centering
    \includegraphics[width=\linewidth,height=0.52\textheight,keepaspectratio]{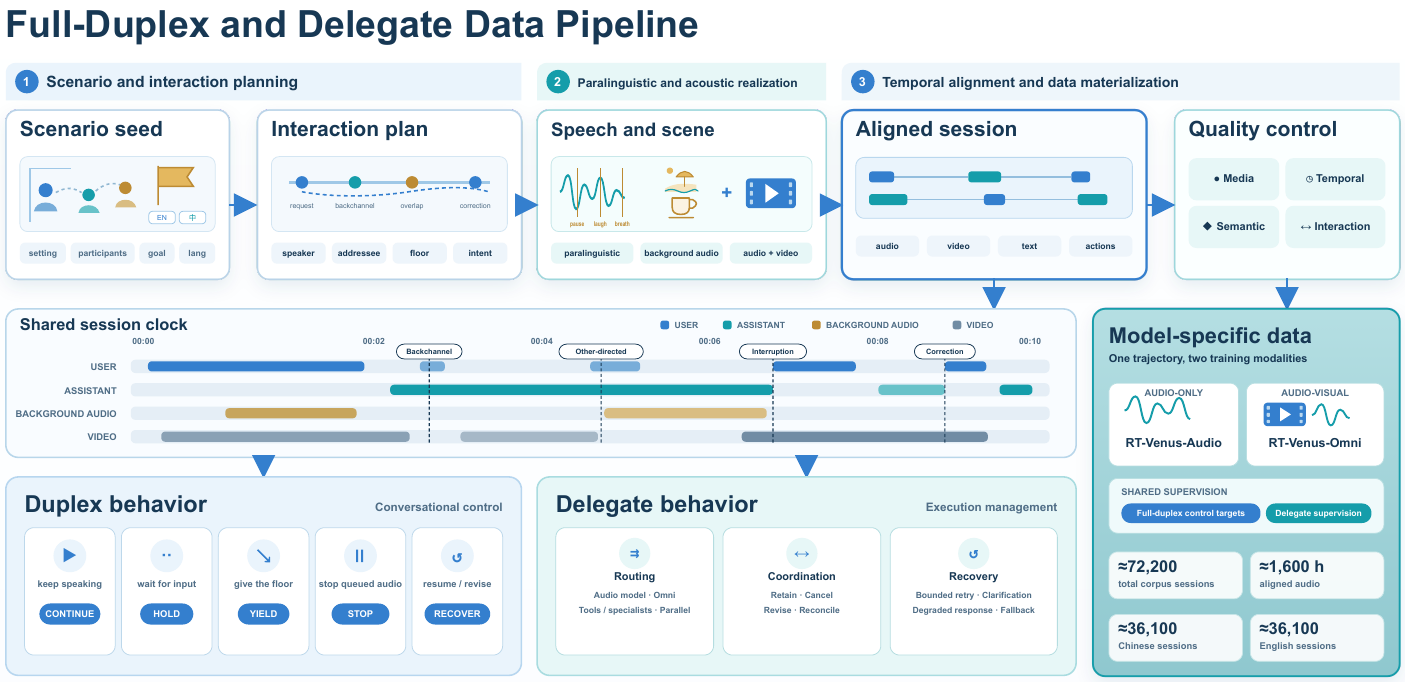}
    \caption{Full-duplex data pipeline.}
    \label{fig:Realtime-Venus-data-pipeline}
\end{figure}

As illustrated in Figure~\ref{fig:Realtime-Venus-data-pipeline}, the pipeline contains three stages: scenario and interaction planning, paralinguistic and acoustic realization, and temporal alignment with model-specific materialization.

\textbf{Scenario and interaction planning.}
Each session begins with a structured scenario seed specifying the setting, participants, user goal, language profile, acoustic environment, and target interaction conditions. The seed is expanded into an event-level plan that determines the semantic progression, speaker and addressee, floor ownership, and whether each incoming event preserves or changes the active user intent. Compositional sampling covers daily communication, information seeking, collaborative tasks, decision making, service interactions, and multi-party conversations. The plans include ordinary turn-taking as well as pauses, backchannels, interruptions, corrections, other-directed speech, and background speech.

\textbf{Paralinguistic and acoustic realization.}
Planned turns are converted into speech and arranged on the session timeline according to their intended boundaries and overlaps. Hesitation, filled pauses, repetition, self-correction, incomplete utterances, laughter, and breathing are added to improve conversational realism and expose cues relevant to turn-taking. Foreground dialogue is mixed with scene-matched secondary speakers and environmental audio, with event onset and duration
varied to produce different overlap conditions. The corpus covers Chinese, English, and code-switching. For audio--visual instances, the composed audio is synchronized with the associated video while retaining the interaction
structure and delegation annotations specified by the plan.

\textbf{Temporal alignment and materialization.}
Audio, video, text, and system actions are mapped to a common monotonic session clock. Each event records its source, temporal span, conversational function, intended addressee, and associated system behavior. The aligned session preserves overlap intervals, response boundaries, and the portion of assistant output affected by an interaction event. It is then materialized as audio-only data for \audiomodel{} or video data for \omnimodel{}. Both modalities share the underlying trajectory and supervision: a compact model-visible vocabulary encodes online interaction-control actions, while non-audible delegation requests,
background results, and execution metadata remain in the structured event stream.

\subsection{Coupling duplex and delegation behaviors}
\label{sec:coupled-supervision}

Each trajectory links an observed event to two complementary targets.
The \textbf{duplex target} specifies the immediate conversational action:
continue speaking, await input, yield the floor, stop queued output,
or revise the response. Selection reflects conversational meaning,
not acoustic overlap alone. Backchannels, other-directed speech,
and unrelated background speech generally preserve the active response,
whereas floor-taking interruptions stop queued audio.
Hesitation may pause responses; intent-changing corrections trigger recovery.

The \textbf{delegation target} specifies computational handling.
Simple or latency-sensitive conversational acts remain with the frontend.
Requests requiring additional perceptual evidence, broader context,
or substantial replanning may invoke a registered capability;
those involving external information or executable actions are routed
to tools or specialized components. Longer operations can run alongside
a cancellable local acknowledgment. Supervision specifies whether
active work is retained, canceled, revised, reconciled with another
result, retried, or replaced by a fallback.

The event's effect on user intent links both targets.
A backchannel preserves speech and computation; other-directed
and background speech create no new route.
A stable-intent interruption may stop playback while retaining
the semantic plan; an intent-changing correction cancels dependent
work and triggers replanning. Tool-dependent requests pair immediate
feedback with asynchronous execution. Each trajectory connects
observable context, interaction control, computation management,
and execution outcomes on a shared timeline.

\section{Evaluation Protocol}
\label{sec:experiments}

We evaluate \systemname{} through complementary assessments of its two frontend models. We assess \omnimodel{} on streaming and offline video understanding. Both \omnimodel{} and \audiomodel{} are evaluated on audio understanding, spoken question answering, and overlap handling on Full-Duplex-Bench v1.5. We also evaluate both frontends within \systemname{} on the tool-use component of Full-Duplex-Bench v3 (FDB-v3) and compare their routing decisions on our internally developed delegate benchmark.

\subsection{Evaluation datasets and metrics}

\textbf{Streaming and offline video understanding.} StreamingBench~\citep{lin2024streaming} covers 18 tasks in real-time visual,
omni-source, and contextual understanding. OVO-Bench~\citep{niu2025ovobench}
covers 12 tasks in real-time visual perception, backward tracing, and
forward active responding. ProactiveVideoQA~\citep{wang2025proactivevideoqa}
reports PAUC across WEB, EGO, TV, and VAD, while
OmniPro~\citep{zhao2026omnipro} reports probe-mode accuracy and online-mode F1.
For offline video understanding, we report overall accuracy on
WorldSense~\citep{hong2026worldsense}, Daily-Omni~\citep{zhou2025dailyomni},
OmniVideoBench~\citep{li2026omnivideobench}, and
LVOmniBench~\citep{tao2026lvomnibench}. The memory study reports accuracy
by duration bin on LVOmniBench, LongVideoBench~\citep{wu2024longvideobench},
and CGBench~\citep{chen2025cgbench}.

\textbf{Audio understanding.} We use MMAU~\citep{sakshi2024mmaumassivemultitaskaudio}, MMAU-Pro~\citep{kumar2025mmau}, MMAR~\citep{ma2025mmar}, and MMSU~\citep{wang2025mmsu} to evaluate acoustic perception and reasoning. MMAU covers speech, music, and sound; MMAU-Pro extends evaluation to challenging long-form, spatial, and multi-audio settings. MMAR emphasizes reasoning from contextual and acoustic evidence, while MMSU assesses spoken-language understanding, including linguistic and paralinguistic cues. We report accuracy under the task configuration used for each benchmark.

\textbf{Spoken question answering.} We evaluate VoiceBench AlpacaEval~\citep{chen2024voicebenchbenchmarkingllmbasedvoice}, Llama Questions~\citep{nachmani2024spectron}, Speech TriviaQA~\citep{defossez2024moshi}, and Speech CMMLU~\citep{ultraevalaudio}. These tasks assess instruction following, factual knowledge, and Chinese academic knowledge through spoken queries. VoiceBench AlpacaEval uses a judge-based response-quality score; the remaining tasks use benchmark-specific measures of answer accuracy.

\textbf{Full-duplex interaction.} Full-Duplex-Bench v1.5~\citep{lin2026fullduplexbench} evaluates user interruption, user backchannel, talking to others, and background speech. Its four response categories are \texttt{C\_RESPOND}, \texttt{C\_RESUME}, \texttt{UNCERTAIN}, and \texttt{UNKNOWN}, denoting responses to overlapping speech, continuation of the preceding response, expressed uncertainty, and irrelevant or absent responses. Responding is the desired behavior for user interruptions; continuing the preceding response is preferred in the other scenarios.

\textbf{Full-duplex tool use.} We evaluate the tool-use component of Full-Duplex-Bench v3 (FDB-v3)~\citep{lin2026fdb_v3}, which tests voice agents on human-recorded speech containing fillers, pauses, hesitations, false starts, and self-corrections. Its tasks span four application domains and require single-step or chained calls to deterministic mock APIs. Tool selection F1 compares expected and predicted tool calls, penalizing missed and extra calls. Argument accuracy measures the semantic correctness of function arguments. Pass@1 is the proportion of episodes in which all expected tools are called, no extra calls are made, and every call has correct arguments. We report these metrics as percentages.

\textbf{Full-duplex delegation.} We evaluate delegation decisions using an internally constructed delegate benchmark. It includes omni and audio-only input modes with different levels of difficulty. Its tasks span three categories: external capabilities, routine interaction, and reasoning. Overall accuracy measures agreement with reference delegation decisions across all requests within each mode. Delegation recall measures the proportion of external-capability requests correctly delegated. Non-delegation specificity measures the proportion of routine requests for which the model correctly avoids delegation. Reasoning accuracy measures correct routing on a mixture of requests requiring delegation and direct handling. We report these metrics as percentages for each input mode.

\subsection{Evaluation settings}

By default, we sample videos at 1 frame per second, retain at most
128 frames, and resize frames to approximately $448\times448$ pixels
while preserving the original aspect ratio. Streaming predictions
are conditioned on observations available at the corresponding query
or response time, while offline tasks permit access to the sampled
evaluation clip before answering. For both \omnimodel{} and \audiomodel{}, input waveforms are converted to mono and resampled to 16~kHz.
During full-duplex evaluation, incoming audio is processed in
one-second chunks, with perception remaining active during speech
generation. When speech output is required, waveforms are generated
at 24~kHz.
We report offline and online baselines separately in the summary
table. In all tables, \textbf{bold} and \underline{underlined} values
indicate the best and second-best distinct scores, respectively.

\FloatBarrier
\subsection{Omni results}

\textbf{Streaming and offline understanding.} Table~\ref{tab:eval_video_summary} compares \omnimodel{} with offline and online baselines across eight benchmarks.

\begin{table}[!htb]
\RealtimeVenusEvalSetup
\renewcommand{\arraystretch}{1.3}
\caption{Streaming and offline video understanding results.}
\label{tab:eval_video_summary}
\setlength{\tabcolsep}{2pt}

\newcommand{\RealtimeVenusVideoHead}[1]{{\fontencoding{T1}\fontfamily{ptm}\fontsize{8}{9.5}\selectfont\bfseries\parbox[c][20pt][c]{\linewidth}{\centering #1}}}

\begin{tabularx}{\linewidth}{@{}>{\raggedright\arraybackslash}p{94pt}>{\centering\arraybackslash}p{17pt}*{8}{>{\centering\arraybackslash}X}@{}}
\toprule
\multirow{2}{*}{\raisebox{-10pt}{\textbf{Model}}}
& \multirow{2}{*}{\raisebox{-10pt}{\textbf{Size}}}
& \multicolumn{4}{c}{\bfseries Streaming benchmarks}
& \multicolumn{4}{c}{\bfseries Offline benchmarks} \\
\cmidrule(lr){3-6}\cmidrule(lr){7-10}
& & \RealtimeVenusVideoHead{Streaming\\Bench}
& \RealtimeVenusVideoHead{OVO-Bench}
& \RealtimeVenusVideoHead{Proactive\\VideoQA}
& \RealtimeVenusVideoHead{OmniPro}
& \RealtimeVenusVideoHead{WorldSense}
& \RealtimeVenusVideoHead{Daily-Omni}
& \RealtimeVenusVideoHead{OmniVideo\\Bench}
& \RealtimeVenusVideoHead{LVOmni\\Bench} \\
\midrule
\multicolumn{10}{l}{\color{black!60}\itshape Offline models:} \\[1pt]
InternVL3.5 & 8B & \RealtimeVenusEvalNum{60.8}{0} & 53.8 & 37.5 & 12.1 & 39.2 & 53.4 & 35.7 & 38.1 \\
Qwen3-VL & 8B & \RealtimeVenusEvalNum{53.5}{0} & 49.8 & 43.4 & 19.5 & 44.0 & 45.6 & 30.7 & 30.2 \\
Qwen3.5 & 9B & \RealtimeVenusEvalNum{57.0}{0} & 55.0 & 47.3 & 22.7 & 45.5 & 47.5 & 31.1 & 32.1 \\
Qwen3-Omni & 30B & \RealtimeVenusEvalNum{64.3}{0} & 61.4 & 44.1 & 22.6 & 49.6 & 72.6 & 38.4 & 39.7 \\
video-SALMONN 2+ & 7B & \RealtimeVenusEvalNum{59.9}{0} & 46.8 & 40.4 & 22.1 & 49.8 & 66.1 & 35.8 & 38.7 \\
AV-Flamingo & 7B & \RealtimeVenusEvalNum{58.5}{0} & 51.5 & 34.6 & 18.6 & 53.2 & 69.4 & 39.9 & 38.0 \\
Gemini-3.5-Flash & \textcolor{black!70}{--} & \RealtimeVenusEvalNum{77.3}{0} & 71.6 & 56.5 & 52.4 & 67.2 & 82.3 & 64.6 & 58.2 \\
\midrule
\addlinespace[3pt]
\multicolumn{10}{l}{\color{black!60}\itshape Online models:} \\[1pt]
LiveStar & 8B & \RealtimeVenusEvalNum{60.2}{0} & 38.0 & 25.0 & 8.3 & 36.1 & 44.7 & 29.3 & 32.5 \\
MMDuet2 & 3B & \RealtimeVenusEvalNum{58.1}{0} & 47.6 & 34.9 & 7.6 & 36.9 & 50.6 & 30.4 & 32.1 \\
JoyAI-VL-Interaction & 8B & \RealtimeVenusEvalNum{63.3}{0} & 56.7 & 39.1 & 18.2 & 43.0 & 54.1 & 36.7 & 36.3 \\
MiniCPM-o 4.5 & 9B & \RealtimeVenusEvalNum{\underline{67.9}}{0} & \underline{60.7} & \textbf{47.5} & \underline{25.6} & \textbf{54.2} & \underline{79.4} & \underline{37.7} & \underline{36.7} \\
\midrule
\rowcolor{blue!5}
\omnimodel{} & 9B & \RealtimeVenusEvalNum{\textbf{70.2}}{0} & \textbf{64.7} & \underline{46.2} & \textbf{29.0} & \underline{54.0} & \textbf{81.3} & \textbf{39.2} & \textbf{38.9} \\
\bottomrule
\end{tabularx}
\RealtimeVenusEvalNote{StreamingBench, OVO-Bench, WorldSense, Daily-Omni, OmniVideoBench, and
LVOmniBench report accuracy. ProactiveVideoQA reports PAUC and OmniPro reports probe-mode accuracy. All scores are shown on a 0--100 scale; offline models are shown for reference.}
\end{table}
\FloatBarrier
The offline baselines are InternVL3.5~\citep{wang2025internvl35}, Qwen3-VL~\citep{bai2025qwen3vl}, Qwen3.5~\citep{qwen2026qwen35}, Qwen3-Omni~\citep{xu2025qwen3omni}, video-SALMONN 2+~\citep{tang2025videosalmonn2}, AV-Flamingo~\citep{ghosh2026avflamingo}, and Gemini-3.5-Flash~\citep{google2026gemini35flash}. The online baselines are LiveStar~\citep{yang2025livestar}, MMDuet2~\citep{wang2025mmduet2}, JoyAI-VL-Interaction~\citep{joyai2026vlinteraction}, and MiniCPM-o 4.5~\citep{cui2026minicpmo45realtimefullduplex}.
\begin{samepage}

\textbf{Results.} \omnimodel{} scores 70.2 on StreamingBench, 64.7 on OVO-Bench, 46.2 on ProactiveVideoQA, and 29.0 on OmniPro. Its offline video understanding scores are 54.0 on WorldSense, 81.3 on Daily-Omni, 39.2 on OmniVideoBench, and 38.9 on LVOmniBench. Compared with MiniCPM-o 4.5, it improves StreamingBench and OVO-Bench accuracy by 2.3 and 4.0 percentage points, respectively, and OmniPro probe-mode accuracy by 3.4 points. It also scores higher on Daily-Omni, OmniVideoBench, and LVOmniBench.
\par\end{samepage}

\textbf{Scope of the comparison.} Among the compared online models, \omnimodel{} achieves the highest scores on six of the eight benchmarks. Its lower scores on ProactiveVideoQA and WorldSense relative to MiniCPM-o 4.5 underscore the need to assess streaming comprehension, proactive response quality, and offline video understanding separately. We report offline baselines separately for reference and restrict the best- and second-best highlighting to online models. These results also do not isolate the contribution of individual training components, which requires controlled ablations.

\begin{samepage}
\textbf{Memory-augmented understanding.} 
Table~\ref{tab:eval-video-memory} reports the performance of the long-video memory module on LVOmniBench, LongVideoBench, and CGBench across different video-duration ranges. 
We report results for both \omnimodel{} and MiniCPM-o 4.5, comparing each backbone with and without memory augmentation. 
Overall, the results show that memory augmentation improves \omnimodel{} across all reported duration bins, with gains for MiniCPM-o 4.5 in several settings as well.

\par\end{samepage}

\begin{table}[!htb]
\RealtimeVenusEvalSetup
\renewcommand{\arraystretch}{1.3}
\centering
\caption{Memory-augmented long-video understanding accuracy (\%). Duration bins are in minutes.}
\label{tab:eval-video-memory}
\setlength{\tabcolsep}{7pt}
\newcommand{\RealtimeVenusMemoryHead}[1]{{\fontencoding{T1}\fontfamily{ptm}\fontsize{8.5}{10}\selectfont\bfseries #1}}

\begin{tabularx}{\linewidth}{
    >{\raggedright\arraybackslash}X
    *{8}{r}
}
\toprule
& \multicolumn{3}{c}{\bfseries LVOmniBench}
& \multicolumn{3}{c}{\bfseries CGBench}
& \multicolumn{2}{c}{\bfseries LongVideoBench} \\
\cmidrule(lr){2-4}
\cmidrule(lr){5-7}
\cmidrule(lr){8-9}

\RealtimeVenusMemoryHead{Model}
& \RealtimeVenusMemoryHead{[10, 30)}
& \RealtimeVenusMemoryHead{[30, 60)}
& \RealtimeVenusMemoryHead{[60, 90)}
& \RealtimeVenusMemoryHead{[40, 50)}
& \RealtimeVenusMemoryHead{[50, 60)}
& \RealtimeVenusMemoryHead{[60, 65)}
& \RealtimeVenusMemoryHead{[15, 40)}
& \RealtimeVenusMemoryHead{[40, 60)} \\
\midrule

MiniCPM-o 4.5
& \underline{38.58} & 38.56 & 32.35
& 34.27 & 36.10 & 34.58
& \underline{52.89} & 55.56 \\

\quad + memory
& \textbf{38.80} & 39.89 & 38.24
& 44.71 & 45.37 & 43.93
& 51.90 & 55.56 \\
\midrule

\rowcolor{blue!5}
\omnimodel{}
& 35.48 & \underline{42.91} & \underline{41.18}
& \underline{47.84} & \underline{46.63} & \underline{49.53}
& \underline{52.89} & \underline{57.14} \\

\rowcolor{blue!5}
\quad + memory
& 36.36 & \textbf{43.10} & \textbf{47.06}
& \textbf{48.71} & \textbf{48.31} & \textbf{50.47}
& \textbf{53.29} & \textbf{61.90} \\

\bottomrule
\end{tabularx}

\end{table}

\FloatBarrier

\textbf{Effect of memory.} 
Memory augmentation consistently improves \omnimodel{} across the evaluated duration ranges. In particular, accuracy increases by 5.88 percentage points on the 60--90-minute subset of LVOmniBench and by 4.76 points on the 40--60-minute subset of LongVideoBench, with gains across all evaluated duration ranges of CGBench as well. For MiniCPM-o 4.5, memory also improves performance in several settings, including improvements of 9.27--10.44 points across CGBench and 5.89 points on the 60--90-minute LVOmniBench subset. The gains are consistent across all evaluated bins for \omnimodel{}, whereas MiniCPM-o 4.5 shows a decrease in the shorter LongVideoBench bin and no change in the longer bin. Thus, the benefit of memory augmentation depends on the backbone and evaluation setting.
\FloatBarrier
\subsection{Speech results}

\textbf{Audio understanding.} Table~\ref{tab:eval-audio-understanding} reports results for both frontend models on MMAU, MMAU-Pro, MMAR, and MMSU. \audiomodel{} achieves accuracies of 78.0\%, 63.2\%, 65.6\%, and 66.0\%, respectively.

The speech comparisons include Fun-Audio-Chat~\citep{tongyi2025funaudiochat}, MiniCPM-o 2.6~\citep{openbmb2025minicpmo26}, Baichuan-Omni-1.5~\citep{li2025baichuanomni15}, Kimi-Audio~\citep{ding2025kimi}, Qwen3-Omni, MiMo-Audio~\citep{xiaomi2025mimoaudio}, Step-Audio2-mini~\citep{wu2025stepaudio2}, and MiniCPM-o 4.5. Model configurations are listed in Tables~\ref{tab:eval-audio-understanding} and~\ref{tab:eval-speech-qa}.

\begin{table}[!htb]
\RealtimeVenusEvalSetup
\renewcommand{\arraystretch}{1.3}
\centering
\caption{Audio understanding accuracy (\%). Higher is better for all benchmarks.}
\label{tab:eval-audio-understanding}
\setlength{\tabcolsep}{2pt}

\begin{tabularx}{\linewidth}{@{}>{\raggedright\arraybackslash}p{110pt}>{\centering\arraybackslash}p{40pt}*{4}{>{\centering\arraybackslash}X}@{}}
\toprule
\textbf{Model} & \RealtimeVenusEvalHead{Size} & \RealtimeVenusEvalHead{MMAU~$\uparrow$} & \RealtimeVenusEvalHead{MMAU-Pro~$\uparrow$} & \RealtimeVenusEvalHead{MMAR~$\uparrow$} & \RealtimeVenusEvalHead{MMSU~$\uparrow$} \\
\midrule
Fun-Audio-Chat & 8B & \RealtimeVenusEvalNum{76.6}{0} & \RealtimeVenusEvalNum{58.0}{0} & 40.7 & \underline{67.8} \\
MiniCPM-o 2.6 & 7B & \RealtimeVenusEvalNum{65.2}{0} & \RealtimeVenusEvalNum{40.5}{0} & 48.6 & 56.5 \\
Baichuan-Omni-1.5 & 7B & \RealtimeVenusEvalNum{65.6}{0} & \RealtimeVenusEvalNum{42.9}{0} & 40.7 & 50.6 \\
Kimi-Audio & 9B & \RealtimeVenusEvalNum{68.4}{0} & \RealtimeVenusEvalNum{56.6}{0} & 60.8 & 59.3 \\
Qwen3-Omni & 30B-A3B & \RealtimeVenusEvalNum{\underline{77.5}}{0} & \RealtimeVenusEvalNum{61.2}{0} & \textbf{66.4} & \textbf{69.0} \\
MiMo-Audio & 7B & \RealtimeVenusEvalNum{74.9}{0} & \RealtimeVenusEvalNum{53.4}{0} & 63.6 & 61.7 \\
Step-Audio2-mini & 7B & \RealtimeVenusEvalNum{68.2}{0} & \RealtimeVenusEvalNum{47.9}{0} & 55.8 & 56.8 \\
MiniCPM-o 4.5 & 9B & \RealtimeVenusEvalNum{76.9}{0} & \RealtimeVenusEvalNum{60.0}{0} & {65.3} & 65.8 \\
\midrule
\rowcolor{blue!5}
\omnimodel{} & 9B & \RealtimeVenusEvalNum{76.9}{0} & \RealtimeVenusEvalNum{\underline{62.0}}{0} & 65.2 & 64.6 \\
\rowcolor{blue!5}
\audiomodel{} & 9B & \RealtimeVenusEvalNum{\textbf{78.0}}{0} & \RealtimeVenusEvalNum{\textbf{63.2}}{0} & \underline{65.6} & 66.0 \\
\bottomrule
\end{tabularx}
\end{table}
\FloatBarrier

\textbf{Task-dependent performance.} \audiomodel{} achieves the highest scores among the compared models on MMAU and MMAU-Pro. On MMAR, it ranks second behind Qwen3-Omni. On MMSU, Qwen3-Omni and Fun-Audio-Chat score higher. These results indicate that the interaction-oriented frontend retains general audio understanding, while contextual acoustic reasoning and fine-grained spoken-language understanding remain areas for improvement.

\textbf{Spoken question answering.} Table~\ref{tab:eval-speech-qa} reports results for both frontend models. \audiomodel{} obtains an AlpacaEval score of 4.81 and accuracies of 83.8\%, 75.7\%, and 67.8\% on Llama Questions, Speech TriviaQA, and Speech CMMLU, respectively.

\begin{table}[!htb]
\RealtimeVenusEvalSetup
\renewcommand{\arraystretch}{1.3}
\centering
\caption{Spoken question-answering results. AlpacaEval reports a judge score; other columns report accuracy (\%). Higher is better.}
\label{tab:eval-speech-qa}
\setlength{\tabcolsep}{2pt}

\begin{tabularx}{\linewidth}{@{}>{\raggedright\arraybackslash}p{110pt}>{\centering\arraybackslash}p{40pt}*{4}{>{\centering\arraybackslash}X}@{}}
\toprule
\textbf{Model} & \RealtimeVenusEvalHead{Size} & \RealtimeVenusEvalHead{AlpacaEval~\ensuremath{\uparrow}} & \RealtimeVenusEvalHead{Llama Q.~\ensuremath{\uparrow}} & \RealtimeVenusEvalHead{TriviaQA~\ensuremath{\uparrow}} & \RealtimeVenusEvalHead{CMMLU~\ensuremath{\uparrow}} \\
\midrule
Fun-Audio-Chat & 8B & \underline{4.80} & 83.3 & 68.1 & \underline{67.7} \\
MiniCPM-o 2.6 & 7B & 4.42 & 78.0 & 51.8 & 51.4 \\
Baichuan-Omni-1.5 & 7B & 4.50 & 78.5 & 63.0 & 58.4 \\
Kimi-Audio & 9B & 4.46 & 79.3 & 62.1 & 67.0 \\
Qwen3-Omni & 30B-A3B & 4.74 & \underline{83.4} & \textbf{75.9} & 47.8 \\
MiMo-Audio & 7B & 4.60 & 79.7 & 52.8 & 56.7 \\
Step-Audio2-mini & 7B & 4.17 & 75.0 & 57.7 & 67.6 \\
MiniCPM-o 4.5 & 9B & \textbf{4.81} & 81.0 & 75.5 & 59.2 \\
\midrule
\rowcolor{blue!5}
\omnimodel{} & 9B & 4.75 & 83.3 & 73.6 & 67.5 \\
\rowcolor{blue!5}
\audiomodel{} & 9B & \textbf{4.81} & \textbf{83.8} & \underline{75.7} & \textbf{67.8} \\
\bottomrule
\end{tabularx}
\end{table}

\textbf{Knowledge coverage.} \audiomodel{} ties MiniCPM-o 4.5 for the highest AlpacaEval score and achieves the highest Llama Questions and Speech CMMLU accuracies among the compared models. It ranks second on Speech TriviaQA, behind Qwen3-Omni. These results cover instruction following and knowledge-based spoken question answering, but do not establish broader reasoning capability. The aggregate scores also do not distinguish errors in speech interpretation from errors in downstream answer generation.

\FloatBarrier
\subsection{Full-duplex results}

\textbf{Overlap handling (v1.5).} Table~\ref{tab:duplex-v15} reports Full-Duplex-Bench v1.5 results for both frontend models. \audiomodel{} has a response rate of 0.75 under user interruption and continuation rates of 0.97, 0.88, and 0.86 under user backchannels, speech directed to others, and background speech, respectively. These scenario-specific results assess whether the model responds to interruptions while maintaining conversational continuity during other overlapping speech.

\begin{table}[!htb]
\RealtimeVenusEvalSetup
\renewcommand{\arraystretch}{1.3}
\caption{Full-Duplex-Bench v1.5 results across four scenarios.}
\label{tab:duplex-v15}
\label{tab:duplex-v15-interruption}
\label{tab:duplex-v15-backchannel}
\label{tab:duplex-v15-others}
\label{tab:duplex-v15-background}
\setlength{\tabcolsep}{2pt}
\newcommand{\RealtimeVenusStateHead}[2]{%
  \mbox{%
    \renewcommand{\ttdefault}{cmtt}%
    \fontsize{8}{9}\mdseries\selectfont
    \texttt{#1}\hspace{0.5pt}$#2$}}

\begin{tabularx}{\linewidth}{@{}>{\raggedright\arraybackslash}p{96pt}*{8}{>{\centering\arraybackslash}X}@{}}
\toprule
 & \multicolumn{2}{c}{\RealtimeVenusEvalHead{User interruption}}
 & \multicolumn{2}{c}{\RealtimeVenusEvalHead{User backchannel}}
 & \multicolumn{2}{c}{\RealtimeVenusEvalHead{Talking to others}}
 & \multicolumn{2}{c}{\RealtimeVenusEvalHead{Background speech}} \\
\cmidrule(lr){2-3}\cmidrule(lr){4-5}
\cmidrule(lr){6-7}\cmidrule(lr){8-9}
\textbf{Model}
 & \RealtimeVenusStateHead{C\_RESPOND}{\uparrow}
 & \RealtimeVenusStateHead{C\_RESUME}{\downarrow}
 & \RealtimeVenusStateHead{C\_RESPOND}{\downarrow}
 & \RealtimeVenusStateHead{C\_RESUME}{\uparrow}
 & \RealtimeVenusStateHead{C\_RESPOND}{\downarrow}
 & \RealtimeVenusStateHead{C\_RESUME}{\uparrow}
 & \RealtimeVenusStateHead{C\_RESPOND}{\downarrow}
 & \RealtimeVenusStateHead{C\_RESUME}{\uparrow} \\
\midrule
Freeze-Omni$^{\dagger}$ & 0.72 & 0.12 & 0.07 & 0.80 & 0.58 & 0.25 & 0.62 & 0.25 \\
Moshi$^{*}$ & 0.50 & 0.26 & 0.02 & 0.06 & 0.20 & 0.19 & 0.21 & 0.07 \\
Gemini 3.1 Live$^{\dagger}$ & 0.77 & 0.20 & 0.02 & 0.95 & 0.27 & 0.66 & 0.28 & 0.66 \\
GPT-4o$^{*}$ & \underline{0.78} & \underline{0.10} & 0.03 & 0.70 & 0.91 & 0.02 & 0.93 & 0.04 \\
Joy-Duplex$^{\dagger}$ & \textbf{0.88} & \textbf{0.07} & \underline{0.01} & \underline{0.96} & 0.17 & 0.72 & \textbf{0.10} & \underline{0.85} \\
MiniCPM-o 4.5 & 0.60 & 0.36 & \textbf{0.00} & 0.95 & 0.18 & 0.79 & 0.16 & 0.82 \\
\midrule
\rowcolor{blue!5}
\omnimodel{} & 0.60 & 0.37 & 0.02 & 0.92 & \textbf{0.06} & \textbf{0.90} & 0.12 & \underline{0.85} \\
\rowcolor{blue!5}
\audiomodel{} & 0.75 & 0.23 & \textbf{0.00} & \textbf{0.97} & \underline{0.11} & \underline{0.88} & \underline{0.11} & \textbf{0.86} \\
\bottomrule
\end{tabularx}

\RealtimeVenusEvalNote{%
Arrows indicate the preferred direction of each metric.
\par
$^{*}$ Results are taken directly from the original Full-Duplex-Bench v1.5 paper~\citep{lin2026fullduplexbench}.
\par
$^{\dagger}$ Results for Joy-Duplex and Gemini 3.1 Live are taken from the JoyAI-Talker paper~\citep{bai2026joyaitalker}.
}
\end{table}

\textbf{Interruption--continuation trade-off.} Table~\ref{tab:duplex-v15} distinguishes overlaps requiring a new response from those requiring continued speech. \audiomodel{} achieves the highest continuation rates for user backchannels and background speech, and ranks second to \omnimodel{} for speech directed to others. However, its interruption-response rate is lower than those of Joy-Duplex, GPT-4o, and Gemini 3.1 Live. Thus, strong continuation does not necessarily imply strong interruption handling. Further evaluation should consider ambiguous overlaps and response-transition timing. We report the four scenarios separately because they require different actions; aggregation would require an explicit weighting scheme.

\textbf{Tool use (FDB-v3).} Table~\ref{tab:voice-tool-results} reports Full-Duplex-Bench v3 results for \systemname{} with both frontends. \omnimodel{} achieves 86.0\% tool selection F1, 53.1\% argument accuracy, and 43.0\% Pass@1, while \audiomodel{} achieves 82.0\%, 52.2\%, and 42.0\%, respectively. The two frontends rank second and fourth, respectively, in tool selection F1. GPT-Realtime leads all three metrics at 87.6\%, 68.0\%, and 60.0\%.

The tool-use baselines comprise the cascaded system, Gemini 2.5 Live~\citep{google2025gemini25nativeaudio}, Gemini 3.1 Live, Grok~\citep{xai2025grokvoice}, Ultravox~\citep{quigley2025ultravox}, GPT-Realtime~\citep{openai2025gptrealtime}, and NemotronLabs VoiceChat-11B~\citep{nvidia2026voicechat}.

\begin{table}[!htb]
\RealtimeVenusEvalSetup
\caption{Full-Duplex-Bench v3 tool-use results (\%). Higher is better for all metrics.}
\label{tab:voice-tool-results}
\setlength{\tabcolsep}{4pt}
\begin{tabularx}{\linewidth}{@{}>{\raggedright\arraybackslash}p{155pt}*{3}{>{\centering\arraybackslash}X}@{}}
\toprule
\textbf{Model} & \RealtimeVenusEvalHead{Tool selection F1~$\uparrow$} & \RealtimeVenusEvalHead{Argument accuracy~$\uparrow$} & \RealtimeVenusEvalHead{Pass@1~$\uparrow$} \\
\midrule
Cascaded & 80.3\% & 56.2\% & 45.0\% \\
Gemini 2.5 Live & 78.6\% & \underline{59.3\%} & 49.0\% \\
Gemini 3.1 Live & 81.7\% & 58.8\% & \underline{54.0\%} \\
Grok & 79.7\% & 54.2\% & 43.0\% \\
Ultravox & 79.4\% & 51.3\% & 41.0\% \\
GPT-Realtime & \textbf{87.6\%} & \textbf{68.0\%} & \textbf{60.0\%} \\
NemotronLabs VoiceChat-11B & 82.5\% & 42.2\% & 33.0\% \\
\midrule
\rowcolor{blue!5}
\omnimodel{} & \underline{86.0\%} & 53.1\% & 43.0\% \\
\rowcolor{blue!5}
\audiomodel{} & {82.0\%} & 52.2\% & 42.0\% \\

\bottomrule
\end{tabularx}
\RealtimeVenusEvalNote{Tool selection denotes F1. Pass@1 requires all expected tool calls, no extra calls, and correct arguments for every call.}
\end{table}

\textbf{Selection versus task completion.} Both frontends remain below GPT-Realtime in argument accuracy and complete-episode success. \audiomodel{} achieves 52.2\% argument accuracy and 42.0\% Pass@1, compared with 53.1\% and 43.0\%, respectively, for \omnimodel{}. These metrics use different success criteria, so their difference cannot be interpreted as an intermediate-stage failure rate. The results motivate closer evaluation of argument grounding, retention of spoken constraints, and coordination across multiple calls. Identifying limiting factors requires examining individual execution trajectories, including failures after appropriate tool selection.

\subsection{Delegate Benchmark}
\label{sec:delegate-benchmark}

\textbf{Benchmark construction.}
We construct an internal delegate benchmark to evaluate whether
real-time interaction models can correctly decide when to delegate.
Given a user request, the model must determine whether to handle
it directly within the conversational frontend or issue a delegation
request to \harnessname{}. The benchmark evaluates delegation accuracy, including both necessary delegation and avoidance of unnecessary delegation.

The benchmark has two input modes with different levels of difficulty:
omni input, which provides audio--visual observations to \omnimodel{},
and audio input, which provides audio-only observations to \audiomodel{}.

\textbf{Task categories.}
The benchmark comprises three categories that probe complementary
aspects of delegation behavior.
\emph{External capabilities} contains requests that require external
capabilities and should therefore trigger delegation.
\emph{Routine interaction} contains requests that the frontend
should handle directly without delegation.
\emph{Reasoning} includes both requests requiring delegation and requests that can be handled directly,
testing whether the model can distinguish reasoning tasks that
require delegation from those suitable for direct handling.
Together, these categories evaluate delegation decisions across
requests with different requirements for external assistance.

\textbf{Evaluation protocol.}
We compare the model's routing decision with the reference label
for each request.
Overall accuracy measures correct routing across the benchmark.
The external capabilities category reports delegation recall, measuring how
reliably the model identifies requests requiring delegation.
The routine interaction category reports non-delegation specificity, measuring
how reliably it avoids unnecessary delegation.
The reasoning category reports routing accuracy within the mixed category.
These metrics jointly characterize delegation decision accuracy
and distinguish missed delegation from unnecessary delegation.

\textbf{Results.}
Table~\ref{tab:delegate-benchmark} reports the performance of
\omnimodel{} and \audiomodel{} on our delegate benchmark.
\omnimodel{} achieves an overall routing accuracy of 75.93\%,
exceeding \audiomodel{} by 7.04 percentage points.
The two variants exhibit different strengths:
\audiomodel{} achieves higher delegation recall in the external
capabilities category (92.22\% versus 68.33\%), whereas \omnimodel{}
achieves higher non-delegation specificity in the routine interaction category
(84.44\% versus 39.44\%).
Both variants achieve 75.00\% routing accuracy in the reasoning category.

\begin{table}[!htb]
\RealtimeVenusEvalSetup
\renewcommand{\arraystretch}{1.3}
\caption{Delegation decision performance on our in-house Delegate Benchmark (\%).}
\label{tab:delegate-benchmark}
\small
\setlength{\tabcolsep}{5pt}

\begin{tabularx}{\linewidth}{@{}>{\raggedright\arraybackslash}Xrrrr@{}}
\toprule
\textbf{Model}
& \textbf{Overall}
& \textbf{External capabilities}
& \textbf{Routine interaction}
& \textbf{Reasoning} \\
\midrule
\omnimodel{}
& \textbf{75.93}
& \underline{68.33}
& \textbf{84.44}
& \textbf{75.00} \\
\audiomodel{}
& \underline{68.89}
& \textbf{92.22}
& \underline{39.44}
& \textbf{75.00} \\
\bottomrule
\end{tabularx}

\RealtimeVenusEvalNote{Overall and reasoning: routing accuracy;
external capabilities: delegation recall;
routine interaction: non-delegation specificity.
Higher is better for all metrics.}
\end{table}

\textbf{Discussion.}
The benchmark identifies different delegation failure patterns in the two models.
\audiomodel{} recognizes most requests requiring external
capabilities but frequently delegates routine requests that
should be handled locally.
\omnimodel{} more reliably avoids unnecessary delegation,
but misses a larger proportion of requests requiring external
capabilities.
These results highlight the need to improve both the recognition
of delegation requirements and direct handling of requests
when delegation is unnecessary.
The benchmark evaluates the correctness of delegation decisions;
successful execution of delegated tasks and integration of their
results into the ongoing conversation require additional evaluation.

\FloatBarrier

\section{Conclusion and Future Work}
\label{sec:conclusion}

Across the evaluated settings, \systemname{} demonstrates
\textbf{competitive multimodal understanding} and strong conversational
continuity under non-interruptive speech. Its asynchronous design
provides \textbf{access to external capabilities while keeping
interaction active}: background reasoning and tool execution proceed
while the frontend continues receiving inputs and managing speech.
Memory augmentation further supports \textbf{hour-scale video
understanding}, with improvements across all evaluated duration bins.
Together, these findings support coordinating immediate conversational
responses with longer-running computation as a promising direction
for assistants that remain engaged with users while handling tasks
beyond the frontend's own capabilities.

Future work will explore \textbf{finer-grained streaming chunks}
to better capture brief events and improve the timing of
conversational responses. We will extend the context window
to support \textbf{longer interactions}, retaining relevant
information and tracking evolving user intent. We will also
investigate \textbf{more complex and diverse tasks} that require
multi-step reasoning, coordinated tool use, and asynchronous
execution, with particular attention to managing concurrent
operations and recovering from delayed or failed tool calls.

\newpage
\beginappendix
\section{Contributions}
\label{sec:contri}

\medskip

\noindent\textbf{Core contributors.}
Ruixiang Zhao\textsuperscript{*}, Hualei Wang\textsuperscript{*},
Renhe Sun\textsuperscript{*}, Enzhi Zhou\textsuperscript{*}, Zihang Liu\textsuperscript{*}, Jincenzi Wu, Xujie Song, Kexin Shi.
{\let\thefootnote\relax\footnotetext{
\textsuperscript{*} Indicates equal contribution. Names marked with an asterisk
are listed first; within the marked and unmarked groups, names are ordered
by last name in reverse alphabetical order.
}}

\medskip
\noindent\textbf{Contributors.}
Pengcheng Zhu, Jiayi Zhou, Baoyue Zhang, Changhao Zhang,
Yuqian Ying, \mbox{Yongxiang Xie}, Zitong Wang, Jinhong Wang,
Tong Niu, Jingjing Liu, Junan Lin,
Haolin He, Hengshuo Chu, Yuhui Chen.

\noindent\textbf{Project leaders.}
Jian Liu\textsuperscript{$\dagger$},
Yuge Huang\textsuperscript{$\dagger$},
Junliang Xing\textsuperscript{$\dagger$},
Yuntao Wang\textsuperscript{$\dagger$}.

\begin{NoHyper}
{%
\let\thefootnote\relax
\footnotetext{%
\textsuperscript{$\dagger$}\ %
\begin{tabular}[t]{@{}l@{}}
Corresponding authors: Jian Liu and Yuge Huang
(\texttt{\{rex.lj, huangyuge.hyg\}@antgroup.com}); \\
Junliang Xing and Yuntao Wang
(\texttt{\{jlxing, yuntaowang\}@tsinghua.edu.cn}).
\end{tabular}
}
}
\end{NoHyper}
\medskip

\noindent\textbf{Project advisor.}
Weiqiang Wang, Chun Yu, Yuanchun Shi.

\input{sections_omni/appendix}
\end{document}

%% file: sections_omni/omni_model.tex
\providecommand{\dtoken}[1]{\mbox{\normalfont\detokenize{#1}}}

\omnimodel{} and \audiomodel{} are the omni-modal and audio-only streaming frontends of \systemname{}, respectively.
Each frontend continuously processes incoming signals, decides whether and when to respond, and generates speech within a single autoregressive interaction loop.
We develop both frontends by adapting MiniCPM-o 4.5~\cite{cui2026minicpmo45realtimefullduplex}, an open-source 9B model. Both incorporate an in-stream delegation protocol that connects the latency-sensitive foreground loop to the Delegate Harness described in Section~\ref{sec:delegate-design}, enabling complex requests to execute asynchronously while perception and speech generation continue. 
\omnimodel{} additionally integrates a training-free long-video memory module that retains relevant visual context across hour-long sessions.

\subsection{Model Architecture} \label{ssec:omni-architecture}

The model family inherits the Omni-Flow architecture of MiniCPM-o 4.5, as shown in Figure~\ref{fig:intrust-omni}. \omnimodel{} encodes aligned visual and audio streams with SigLIP2~\cite{siglip2} and Whisper-Medium~\cite{Whisper}, whereas \audiomodel{} removes the ViT-based visual branch and processes only streaming audio. The projected features are consumed by a Qwen3-8B language backbone, whose generated text and hidden states condition discrete S3 speech-token prediction. A streaming flow-matching decoder converts these tokens into waveform chunks using reference audio from the system prompt~\cite{du2024cosyvoice2}.

Streaming is organized into one-second units.
Each \omnimodel{} unit interleaves visual tokens from the current frame with temporally aligned audio features, while each \audiomodel{} unit contains only audio features. 
At each unit, the language model predicts \dtoken{<|listen|>} or \dtoken{<|speak|>}.
For speaking units, response text is generated and used to condition aligned S3 speech-token generation.
The token \dtoken{<|chunk_eos|>} closes a speech chunk, while \dtoken{<|turn_eos|>} marks the end of an assistant turn.
This chunk-wise schedule keeps perception concurrent with speaking and the
unspoken continuation revisable, supporting two interaction capabilities:
\omnimodel{} can respond \emph{omni-proactively}---watching and listening
continuously and initiating speech when a new event warrants it---while both
models conduct \emph{full-duplex conversation}, distinguishing backchannels
from interruptions to stop, repair, or redirect the unspoken continuation.
These capabilities are supervised by the proactive and full-duplex data in Section~\ref{ssec:omni-data}.

\subsection{Training-Free Long-Video Memory} \label{ssec:omni-memory}
Streaming audio-visual input grows continuously, while the base model can maintain only a bounded rolling context window. 
As a session extends to tens of minutes or even hours, earlier content gradually falls outside the window, making historical information inaccessible to subsequent queries. 
We therefore attach an external long-term memory module to the streaming pipeline. 
The module requires no additional training or parameter updates and remains decoupled from the fine-tuned dialogue policy.

\textbf{Memory construction.} 
Storing the visual representation of every sampled frame would cause the memory size and subsequent retrieval cost to grow continuously over time. 
Inspired by the predictive visual coding criterion of AdaCodec~\cite{hou2026adacodec}, we introduce a visual memory gating mechanism based on motion-compensated prediction cost. 
Each sampled frame is compared with the preceding sampled frame through lightweight block-level motion matching, and the resulting prediction residual and motion cost are used to estimate visual changes between them. 
To reduce the risk of discarding short-lived visual events, we further introduce local-change protection and a maximum consecutive-dropping constraint. 
The gating mechanism only determines whether a sampled frame is archived into long-term memory and does not affect its normal processing within the short-term context. 
Retained frames are stored together with their visual representations and timestamps, while audio is preserved independently and aligned with the video timeline.

\textbf{Memory retrieval.} 
Inspired by the MaxSim operator~\cite{khattab2020colbert,wu2026semantic}, the module performs fine-grained matching between query tokens and the visual tokens of each stored historical frame. For each query token, we take its maximum cosine similarity over the visual tokens within a frame as its token-level matching score. To emphasize query tokens whose matching scores vary more across historical frames, we compute the median absolute deviation (MAD) of these scores and normalize the resulting values into token weights. The weighted token-level scores are then aggregated to obtain the semantic relevance score of each frame.
Based on the semantic relevance scores, we first identify a set of candidate frames and rank them by relevance. For each candidate, we compare it with higher-ranked candidates to estimate its visual novelty. Inspired by the relevance-diversity principle of Maximal Marginal Relevance (MMR), we combine semantic relevance and visual novelty into a final retrieval score. These scores are computed once based on the initial relevance ranking and remain fixed during selection. The highest-scoring candidates are selected as historical evidence, aiming to preserve query-relevant information while reducing redundant visual content.

\textbf{Context reassembly.} 
For each retrieved frame, the module retrieves its temporally adjacent audio segments and merges overlapping temporal intervals. The selected historical frames and their associated audio are then arranged chronologically together with the recent short-term audio-visual window and the current query to form the context for answer generation. In this way, the model can recover relevant historical information beyond its rolling context window without continuously retaining the complete input history.

%% file: sections_omni/appendix.tex
\providecommand{\dtoken}[1]{\mbox{\normalfont\detokenize{#1}}}

\section{Training Data Example}
\label{app:training-data-example}

We present one training example for each of the two data families central to this report: proactive duplex interaction and delegate workflows. Both are rendered from raw training records. In the unit-level format of Section~\ref{ssec:omni-architecture}, each \dtoken{<unit>} contains one second of aligned visual (\dtoken{<image>}) and audio (\dtoken{<audio>}) input. The assistant responds with \dtoken{<|listen|>} for silent perception or \dtoken{<|speak|>} followed by chunked text ending with \dtoken{<|chunk_eos|>}; \dtoken{<|turn_eos|>} marks the end of a turn. Long runs of identical \dtoken{<|listen|>} units are omitted.

\subsection{Omni-Proactive Example}
\label{app:example-proactive}

The assistant receives an instruction at the beginning of the stream and
must fire the instant the requested event occurs, staying silent both before and after.

\begin{tcolorbox}[
    breakable,
    colback = gray!6, colframe = black!60, boxrule = 0.6pt,
    rounded corners, left = 4pt, right = 4pt, top = 2pt, bottom = 2pt,
    fontupper = {\ttfamily\small}, halign = left
]
\{"role": "user", "content": "<unit>\textbackslash n<image>\textbackslash n<audio>\textbackslash n
Let me know when the commentator reacts to a wicket."\}\\[2pt]
\{"role": "assistant", "content": "<|listen|>"\}\\[2pt]
$\;\;\vdots$ (4 consecutive listen units omitted)\\[2pt]
\{"role": "user", "content": "</unit><unit>\textbackslash n<image>\textbackslash n<audio>"\}\\[2pt]
\{"role": "assistant", "content": "<|speak|>The commentator just reacted<|chunk\_eos|>"\}\\[2pt]
\{"role": "user", "content": "</unit><unit>\textbackslash n<image>\textbackslash n<audio>"\}\\[2pt]
\{"role": "assistant", "content": "<|speak|> to the wicket<|chunk\_eos|>"\}\\[2pt]
\{"role": "user", "content": "</unit><unit>\textbackslash n<image>\textbackslash n<audio>"\}\\[2pt]
\{"role": "assistant", "content": "<|speak|>, shouting 'Oh<|chunk\_eos|>"\}\\[2pt]
\{"role": "user", "content": "</unit><unit>\textbackslash n<image>\textbackslash n<audio>"\}\\[2pt]
\{"role": "assistant", "content": "<|speak|>, he's got<|chunk\_eos|>"\}\\[2pt]
\{"role": "user", "content": "</unit><unit>\textbackslash n<image>\textbackslash n<audio>"\}\\[2pt]
\{"role": "assistant", "content": "<|speak|> him!'<|chunk\_eos|>"\}\\[2pt]
\{"role": "user", "content": "</unit><unit>\textbackslash n<image>\textbackslash n<audio>"\}\\[2pt]
\{"role": "assistant", "content": "<|speak|><|turn\_eos|><|chunk\_eos|>"\}\\[2pt]
$\;\;\vdots$ (7 trailing listen units omitted)
\end{tcolorbox}

The example supervises the proactive response capability of \omnimodel{}: the
model must remain silent while nothing relevant occurs and speak immediately
when the triggering phrase appears in the audio stream.

\subsection{Delegate Example}
\label{app:example-delegate}

The example shows the full delegate cycle of Section~\ref{sec:delegate-design}
within a continuous video session.

\begin{tcolorbox}[
    breakable,
    colback = gray!6, colframe = black!60, boxrule = 0.6pt,
    rounded corners, left = 4pt, right = 4pt, top = 2pt, bottom = 2pt,
    fontupper = {\ttfamily\small}, halign = left
]
\{"role": "user", "content": "<unit>\textbackslash n<image>\textbackslash n<audio>"\}\\[2pt]
\{"role": "assistant", "content": "<|listen|>"\}\\[2pt]
$\;\;\vdots$ (39 consecutive listen units omitted)\\[2pt]
\{"role": "user", "content": "</unit><unit>\textbackslash n<image>\textbackslash n<audio>\textbackslash n
\begin{CJK}{UTF8}{gkai}今天北京\begin{CJK}{UTF8}{gkai}限行\end{CJK}\begin{CJK}{UTF8}{gkai}尾号是啥，我这京牌车尾号7能开吗？\end{CJK}\end{CJK}"\}\\[2pt]
\{"role": "assistant", "content": "<|speak|>\begin{CJK}{UTF8}{gkai}好，\end{CJK}<|chunk\_eos|>"\}\\[2pt]
\{"role": "user", "content": "</unit><unit>\textbackslash n<image>\textbackslash n<audio>"\}\\[2pt]
\{"role": "assistant", "content": "<|speak|>\begin{CJK}{UTF8}{gkai}我查\end{CJK}<|chunk\_eos|>"\}\\[2pt]
\{"role": "user", "content": "</unit><unit>\textbackslash n<image>\textbackslash n<audio>"\}\\[2pt]
\{"role": "assistant", "content": "<|speak|>\begin{CJK}{UTF8}{gkai}一下今天的\end{CJK}<|chunk\_eos|>"\}\\[2pt]
\{"role": "user", "content": "</unit><unit>\textbackslash n<image>\textbackslash n<audio>"\}\\[2pt]
\{"role": "assistant", "content": "<|speak|>\begin{CJK}{UTF8}{gkai}限行\end{CJK}<|chunk\_eos|>"\}\\[2pt]
\{"role": "user", "content": "</unit><unit>\textbackslash n<image>\textbackslash n<audio>"\}\\[2pt]
\{"role": "assistant", "content": "<|speak|>\begin{CJK}{UTF8}{gkai}规则。\end{CJK}<|chunk\_eos|>"\}\\[2pt]
\{"role": "user", "content": "</unit><unit>\textbackslash n<image>\textbackslash n<audio>"\}\\[2pt]
\{"role": "assistant", "content": "<|speak|><delegate>\begin{CJK}{UTF8}{gkai}查询今天北京\begin{CJK}{UTF8}{gkai}限行\end{CJK}\begin{CJK}{UTF8}{gkai}尾号及京牌尾号7能否通行。\end{CJK}\end{CJK}</delegate><|chunk\_eos|>"\}\\[2pt]
\{"role": "user", "content": "</unit><unit>\textbackslash n<image>\textbackslash n<audio>"\}\\[2pt]
\{"role": "assistant", "content": "<|speak|><|turn\_eos|><|chunk\_eos|>"\}\\[2pt]
\{"role": "user", "content": "</unit><unit>\textbackslash n<image>\textbackslash n<audio>"\}\\[2pt]
\{"role": "assistant", "content": "<|listen|>"\}\\[2pt]
\{"role": "user", "content": "</unit><unit>\textbackslash n<image>\textbackslash n<audio>"\}\\[2pt]
\{"role": "assistant", "content": "<|listen|>"\}\\[2pt]
\{"role": "user", "content": "</unit><unit>\textbackslash n<image>\textbackslash n<audio>"\}\\[2pt]
\{"role": "assistant", "content": "<|listen|>"\}\\[2pt]
\{"role": "user", "content": "</unit><unit>\textbackslash n<image>\textbackslash n<audio>\textbackslash n<backend>\begin{CJK}{UTF8}{gkai}你刚才问的\end{CJK}\begin{CJK}{UTF8}{gkai}北京\end{CJK}\begin{CJK}{UTF8}{gkai}限行\end{CJK}\begin{CJK}{UTF8}{gkai}查到了：今天\end{CJK}\begin{CJK}{UTF8}{gkai}限行\end{CJK}\begin{CJK}{UTF8}{gkai}尾号为\end{CJK}\begin{CJK}{UTF8}{gkai}2和7，\end{CJK}\begin{CJK}{UTF8}{gkai}你的京牌尾\end{CJK}\begin{CJK}{UTF8}{gkai}号7在五\end{CJK}\begin{CJK}{UTF8}{gkai}环路以内道路\end{CJK}\begin{CJK}{UTF8}{gkai}限行\end{CJK}\begin{CJK}{UTF8}{gkai}，时段\end{CJK}\begin{CJK}{UTF8}{gkai}是7点到\end{CJK}\begin{CJK}{UTF8}{gkai}20点。\end{CJK}</backend>"\}\\[2pt]
\{"role": "assistant", "content": "<|speak|>\begin{CJK}{UTF8}{gkai}你刚才问的\end{CJK}<|chunk\_eos|>"\}\\[2pt]
\{"role": "user", "content": "</unit><unit>\textbackslash n<image>\textbackslash n<audio>"\}\\[2pt]
\{"role": "assistant", "content": "<|speak|>\begin{CJK}{UTF8}{gkai}北京\end{CJK}\begin{CJK}{UTF8}{gkai}限行\end{CJK}\begin{CJK}{UTF8}{gkai}查\end{CJK}<|chunk\_eos|>"\}\\[2pt]
\{"role": "user", "content": "</unit><unit>\textbackslash n<image>\textbackslash n<audio>"\}\\[2pt]
\{"role": "assistant", "content": "<|speak|>\begin{CJK}{UTF8}{gkai}到了：今天限\end{CJK}<|chunk\_eos|>"\}\\[2pt]
\{"role": "user", "content": "</unit><unit>\textbackslash n<image>\textbackslash n<audio>"\}\\[2pt]
\{"role": "assistant", "content": "<|speak|>\begin{CJK}{UTF8}{gkai}行尾号为\end{CJK}<|chunk\_eos|>"\}\\[2pt]
\{"role": "user", "content": "</unit><unit>\textbackslash n<image>\textbackslash n<audio>"\}\\[2pt]
\{"role": "assistant", "content": "<|speak|>\begin{CJK}{UTF8}{gkai}2和7，\end{CJK}<|chunk\_eos|>"\}\\[2pt]
\{"role": "user", "content": "</unit><unit>\textbackslash n<image>\textbackslash n<audio>"\}\\[2pt]
\{"role": "assistant", "content": "<|speak|>\begin{CJK}{UTF8}{gkai}你的京牌尾\end{CJK}<|chunk\_eos|>"\}\\[2pt]
\{"role": "user", "content": "</unit><unit>\textbackslash n<image>\textbackslash n<audio>"\}\\[2pt]
\{"role": "assistant", "content": "<|speak|>\begin{CJK}{UTF8}{gkai}号7在五\end{CJK}<|chunk\_eos|>"\}\\[2pt]
\{"role": "user", "content": "</unit><unit>\textbackslash n<image>\textbackslash n<audio>"\}\\[2pt]
\{"role": "assistant", "content": "<|speak|>\begin{CJK}{UTF8}{gkai}环路以内道路\end{CJK}<|chunk\_eos|>"\}\\[2pt]
\{"role": "user", "content": "</unit><unit>\textbackslash n<image>\textbackslash n<audio>"\}\\[2pt]
\{"role": "assistant", "content": "<|speak|>\begin{CJK}{UTF8}{gkai}限行\end{CJK}\begin{CJK}{UTF8}{gkai}，时段\end{CJK}<|chunk\_eos|>"\}\\[2pt]
\{"role": "user", "content": "</unit><unit>\textbackslash n<image>\textbackslash n<audio>"\}\\[2pt]
\{"role": "assistant", "content": "<|speak|>\begin{CJK}{UTF8}{gkai}是7点到\end{CJK}<|chunk\_eos|>"\}\\[2pt]
\{"role": "user", "content": "</unit><unit>\textbackslash n<image>\textbackslash n<audio>"\}\\[2pt]
\{"role": "assistant", "content": "<|speak|>\begin{CJK}{UTF8}{gkai}20点。\end{CJK}<|chunk\_eos|>"\}\\[2pt]
\{"role": "user", "content": "</unit><unit>\textbackslash n<image>\textbackslash n<audio>"\}\\[2pt]
\{"role": "assistant", "content": "<|speak|><|turn\_eos|><|chunk\_eos|>"\}\\[2pt]
$\;\;\vdots$ (trailing listen units omitted)
\end{tcolorbox}

The first response chunks acknowledge the spoken query and commit to looking
it up; the \dtoken{<delegate>} span dispatches the task to the Harness and the
turn closes with \dtoken{<|turn_eos|>}, so the frontend returns to listening
while the backend executes. When the \dtoken{<backend>} result re-enters the
session, the model re-presents it in chunked spoken form within the same
conversational timeline.

%% file: paper.bbl
\newpage

\begin{thebibliography}{64}
\providecommand{\natexlab}[1]{#1}
\providecommand{\url}[1]{\texttt{#1}}
\expandafter\ifx\csname urlstyle\endcsname\relax
  \providecommand{\doi}[1]{doi: #1}\else
  \providecommand{\doi}{doi: \begingroup \urlstyle{rm}\Url}\fi

\bibitem[Bai et~al.(2025)Bai, Cai, Chen, Chen, Chen, Cheng, Deng, Ding, Gao, Ge, Ge, Guo, Huang, Huang, Huang, Hui, Jiang, Li, Li, Li, Li, Lin, Lin, Liu, Liu, Liu, Liu, Liu, Liu, Lu, Luo, Lv, Men, Meng, Ren, Ren, Song, Sun, Tang, Tu, Wan, Wang, Wang, Wang, Wang, Xie, Xu, Xu, Xu, Yang, Yang, Yang, Yang, Yu, Zhang, Zhang, Zhang, Zheng, Zhong, Zhou, Zhou, Zhou, Zhu, and Zhu]{bai2025qwen3vl}
Shuai Bai, Yuxuan Cai, Ruizhe Chen, et~al.
\newblock {Qwen3-VL} technical report.
\newblock \emph{arXiv preprint arXiv:2511.21631}, 2025.
\newblock \doi{10.48550/arXiv.2511.21631}.
\newblock URL \url{https://arxiv.org/abs/2511.21631}.

\bibitem[Bai et~al.(2026)Bai, Chen, Chen, Deng, Dong, Duan, Gu, Han, Huang, Ke, Li, Li, Liang, Liu, Liu, Miao, Wang, Wang, Wang, Wang, Wang, Xiao, Xue, Xue, Yu, Zhang, Zhang, Zhang, and Zhu]{bai2026joyaitalker}
Yinhao Bai, Jinming Chen, Yafeng Chen, et~al.
\newblock {JoyAI-Talker}: Full-duplex speech interactive large model built for empathetic voice agents.
\newblock \emph{arXiv preprint arXiv:2608.01119}, 2026.
\newblock \doi{10.48550/arXiv.2608.01119}.
\newblock URL \url{https://arxiv.org/abs/2608.01119}.

\bibitem[Chen et~al.(2025)Chen, Liu, Huang, He, Pei, Xu, Wang, Lu, and Wang]{chen2025cgbench}
Guo Chen, Yicheng Liu, Yifei Huang, et~al.
\newblock {CG-Bench}: Clue-grounded question answering benchmark for long video understanding.
\newblock In \emph{The Thirteenth International Conference on Learning Representations}, 2025.
\newblock URL \url{https://openreview.net/forum?id=le4IoZZHy1}.

\bibitem[Chen et~al.(2024)Chen, Yue, Zhang, Gao, Tan, and Li]{chen2024voicebenchbenchmarkingllmbasedvoice}
Yiming Chen, Xianghu Yue, Chen Zhang, et~al.
\newblock {VoiceBench}: Benchmarking {LLM}-based voice assistants, 2024.
\newblock URL \url{https://arxiv.org/abs/2410.17196}.

\bibitem[Chien et~al.(2026)Chien, Orsini, Kharitonov, Zeghidour, Livescu, and D{\'e}fossez]{chien2026moshirag}
Chung-Ming Chien, Manu Orsini, Eugene Kharitonov, et~al.
\newblock {MoshiRAG}: Asynchronous knowledge retrieval for full-duplex speech language models, 2026.
\newblock URL \url{https://arxiv.org/abs/2604.12928}.

\bibitem[Chu et~al.(2023)Chu, Xu, Zhou, Yang, Zhang, Yan, Zhou, and Zhou]{chu2023qwenaudio}
Yunfei Chu, Jin Xu, Xiaohuan Zhou, et~al.
\newblock {Qwen-Audio}: Advancing universal audio understanding via unified large-scale audio-language models, 2023.
\newblock URL \url{https://arxiv.org/abs/2311.07919}.

\bibitem[Chu et~al.(2024)Chu, Xu, Yang, Wei, Wei, Guo, Leng, Lv, He, Lin, Zhou, and Zhou]{chu2024qwen2audio}
Yunfei Chu, Jin Xu, Qian Yang, et~al.
\newblock {Qwen2-Audio} technical report, 2024.
\newblock URL \url{https://arxiv.org/abs/2407.10759}.

\bibitem[Cui et~al.(2026)Cui, Xu, Wang, Yu, Sun, Xu, Wang, He, Ma, Cai, et~al.]{cui2026minicpmo45realtimefullduplex}
Junbo Cui, Bokai Xu, Chongyi Wang, et~al.
\newblock {MiniCPM-o} 4.5: Towards real-time full-duplex omni-modal interaction.
\newblock \emph{arXiv preprint arXiv:2604.27393}, 2026.
\newblock URL \url{https://arxiv.org/abs/2604.27393}.

\bibitem[D\'efossez et~al.(2024)D\'efossez, Mazar\'e, Orsini, Royer, P\'erez, J\'egou, Grave, and Zeghidour]{defossez2024moshi}
Alexandre D\'efossez, Laurent Mazar\'e, Manu Orsini, et~al.
\newblock {Moshi}: A speech-text foundation model for real-time dialogue.
\newblock \emph{arXiv preprint arXiv:2410.00037}, 2024.
\newblock URL \url{https://arxiv.org/abs/2410.00037}.

\bibitem[Ding et~al.(2025)Ding, Ju, Leng, Liu, Liu, Shang, Shen, Song, Tan, Tang, et~al.]{ding2025kimi}
Ding Ding, Zeqian Ju, Yichong Leng, et~al.
\newblock {Kimi-Audio} technical report.
\newblock \emph{arXiv preprint arXiv:2504.18425}, 2025.

\bibitem[Dosovitskiy et~al.(2020)Dosovitskiy, Beyer, Kolesnikov, Weissenborn, Zhai, Unterthiner, Dehghani, Minderer, Heigold, Gelly, Uszkoreit, and Houlsby]{dosovitskiy2020image}
Alexey Dosovitskiy, Lucas Beyer, Alexander Kolesnikov, et~al.
\newblock An image is worth 16x16 words: Transformers for image recognition at scale.
\newblock \emph{arXiv preprint arXiv:2010.11929}, 2020.

\bibitem[Du et~al.(2024{\natexlab{a}})Du, Chen, Zhang, Hu, Lu, Yang, Hu, Zheng, Gu, Ma, et~al.]{du2024cosyvoice}
Zhihao Du, Qian Chen, Shiliang Zhang, et~al.
\newblock {CosyVoice}: A scalable multilingual zero-shot text-to-speech synthesizer based on supervised semantic tokens.
\newblock \emph{arXiv preprint arXiv:2407.05407}, 2024{\natexlab{a}}.

\bibitem[Du et~al.(2024{\natexlab{b}})Du, Wang, Chen, Shi, Lv, Zhao, Gao, Yang, Gao, Wang, et~al.]{du2024cosyvoice2}
Zhihao Du, Yuxuan Wang, Qian Chen, et~al.
\newblock {CosyVoice} 2: Scalable streaming speech synthesis with large language models.
\newblock \emph{arXiv preprint arXiv:2412.10117}, 2024{\natexlab{b}}.

\bibitem[{Gemini Team, Google}(2023)]{geminiteam2023gemini}
{Gemini Team, Google}.
\newblock {Gemini}: A family of highly capable multimodal models, 2023.
\newblock URL \url{https://arxiv.org/abs/2312.11805}.

\bibitem[Ghosh et~al.(2026)Ghosh, Goel, Jayakumar, Koroshinadze, Anand, Gururani, Ye, Biswas, Su, Hosseini-Asl, gil Lee, Kong, Kim, Kim, Sakshi, Duraiswami, Manocha, Tao, Shoeybi, Catanzaro, Liu, and Ping]{ghosh2026avflamingo}
Sreyan Ghosh, Arushi Goel, Kaousheik Jayakumar, et~al.
\newblock {Audio-Visual Flamingo}: Open audio-visual intelligence for long and complex videos.
\newblock \emph{arXiv preprint arXiv:2607.16107}, 2026.
\newblock \doi{10.48550/arXiv.2607.16107}.
\newblock URL \url{https://arxiv.org/abs/2607.16107}.

\bibitem[{Google}(2025)]{google2025gemini25nativeaudio}
{Google}.
\newblock {Gemini} 2.5 native audio upgrade, plus text-to-speech model updates, December 2025.
\newblock URL \url{https://blog.google/products-and-platforms/products/gemini/gemini-audio-model-updates/}.

\bibitem[{Google DeepMind}(2026)]{google2026gemini35flash}
{Google DeepMind}.
\newblock {Gemini 3.5 Flash} model card, May 2026.
\newblock URL \url{https://deepmind.google/models/model-cards/gemini-3-5-flash/}.

\bibitem[Hong et~al.(2026)Hong, Yan, Cai, Jiang, Hu, and Xie]{hong2026worldsense}
Jack Hong, Shilin Yan, Jiayin Cai, et~al.
\newblock {WorldSense}: Evaluating real-world omnimodal understanding for multimodal {LLMs}.
\newblock In \emph{The Fourteenth International Conference on Learning Representations}, 2026.
\newblock URL \url{https://arxiv.org/abs/2502.04326}.

\bibitem[Hou et~al.(2026)Hou, Huang, Liang, Si, Li, Dong, Shao, Li, Wang, Duan, et~al.]{hou2026adacodec}
Haowen Hou, Zhen Huang, Zheming Liang, et~al.
\newblock Adacodec: A predictive visual code for video mllms.
\newblock \emph{arXiv preprint arXiv:2606.02569}, 2026.

\bibitem[Huang et~al.(2026)Huang, Zhang, Yu, Shi, Ma, Xu, Gao, Hao, He, and Liu]{huang2026duplexomni}
Muye Huang, Lingling Zhang, Xingyu Yu, et~al.
\newblock {DuplexOmni}: Real-time listening, seeing, thinking, and speaking for full-duplex interaction.
\newblock \emph{arXiv preprint arXiv:2606.09186}, 2026.
\newblock \doi{10.48550/arXiv.2606.09186}.
\newblock URL \url{https://arxiv.org/abs/2606.09186}.

\bibitem[Khattab and Zaharia(2020)]{khattab2020colbert}
Omar Khattab and Matei Zaharia.
\newblock Colbert: Efficient and effective passage search via contextualized late interaction over bert.
\newblock In \emph{Proceedings of the 43rd International ACM SIGIR conference on research and development in Information Retrieval}, pages 39--48, 2020.

\bibitem[Kumar et~al.(2025)Kumar, Sedl{\'a}{\v{c}}ek, Lokegaonkar, L{\'o}pez, Yu, Anand, Ryu, Chen, Pli{\v{c}}ka, Hlav{\'a}{\v{c}}ek, Ellingwood, Udupa, Hou, Ferner, Barahona, Bola{\~n}os, Rahi, Herrera-Alarc{\'o}n, Dixit, Patil, Deshmukh, Koroshinadze, Liu, Perera, Zanou, Stafylakis, Chung, Harwath, Zhang, Manocha, Lozano-Diez, Kesiraju, Ghosh, and Duraiswami]{kumar2025mmau}
Sonal Kumar, {\v{S}}imon Sedl{\'a}{\v{c}}ek, Vaibhavi Lokegaonkar, et~al.
\newblock {MMAU-Pro}: A challenging and comprehensive benchmark for holistic evaluation of audio general intelligence.
\newblock \emph{arXiv preprint arXiv:2508.13992}, 2025.
\newblock URL \url{https://arxiv.org/abs/2508.13992}.

\bibitem[Li et~al.(2026)Li, Chen, Ji, Xu, Cui, Li, Zhang, Wang, Song, Zhang, He, Liu, Wang, Wang, Tang, Wu, Luo, Pan, Xie, Zhang, Wang, Tian, Wang, Cao, Dai, Wang, Wen, Ma, Pan, Chang, Taheri, Xia, Plachouras, Benetos, Li, Zhang, Yang, Peng, Wang, Liu, Peng, Zhang, and Liu]{li2026omnivideobench}
Caorui Li, Yu~Chen, Yiyan Ji, et~al.
\newblock {OmniVideoBench}: Towards audio-visual understanding evaluation for omni {MLLMs}.
\newblock In \emph{The Fourteenth International Conference on Learning Representations}, 2026.
\newblock URL \url{https://openreview.net/forum?id=ItRYEe8E61}.

\bibitem[Li et~al.(2025)Li, Liu, Zhang, Zhang, Chen, Li, Li, Liu, Ming, Dong, et~al.]{li2025baichuanomni15}
Yadong Li, Jun Liu, Tao Zhang, et~al.
\newblock {Baichuan-Omni-1.5} technical report.
\newblock \emph{arXiv preprint arXiv:2501.15368}, 2025.
\newblock \doi{10.48550/arXiv.2501.15368}.
\newblock URL \url{https://arxiv.org/abs/2501.15368}.

\bibitem[Lin et~al.(2026{\natexlab{a}})Lin, Chen, Chen, and Lee]{lin2026fdb_v3}
Guan-Ting Lin, Chen Chen, Zhehuai Chen, et~al.
\newblock {Full-Duplex-Bench-v3}: Benchmarking tool use for full-duplex voice agents under real-world disfluency.
\newblock \emph{arXiv preprint arXiv:2604.04847}, 2026{\natexlab{a}}.
\newblock URL \url{https://arxiv.org/abs/2604.04847}.

\bibitem[Lin et~al.(2026{\natexlab{b}})Lin, Kuan, Wang, Lian, Li, Watanabe, and Lee]{lin2026fullduplexbench}
Guan-Ting Lin, Shih-Yun~Shan Kuan, Qirui Wang, et~al.
\newblock {Full-Duplex-Bench v1.5}: Evaluating overlap handling for full-duplex speech models.
\newblock In \emph{ICASSP 2026 - 2026 IEEE International Conference on Acoustics, Speech and Signal Processing (ICASSP)}, 2026{\natexlab{b}}.
\newblock URL \url{https://arxiv.org/abs/2507.23159}.

\bibitem[Lin et~al.(2024)Lin, Fang, Chen, Wan, Luo, Li, Liu, and Sun]{lin2024streaming}
Junming Lin, Zheng Fang, Chi Chen, et~al.
\newblock {StreamingBench}: Assessing the gap for {MLLMs} to achieve streaming video understanding.
\newblock \emph{arXiv preprint arXiv:2411.03628}, 2024.
\newblock URL \url{https://arxiv.org/abs/2411.03628}.

\bibitem[Ma et~al.(2025)Ma, Ma, Zhu, Yang, Chao, Xu, Chen, Chen, Chen, Cong, Li, Li, Li, Li, Li, Lian, Liang, Liu, Niu, Wang, Wang, Wang, Wu, Yang, Yu, Yuan, Zheng, Zhou, Zhu, Xue, Benetos, Yu, Chng, and Chen]{ma2025mmar}
Ziyang Ma, Yinghao Ma, Yanqiao Zhu, et~al.
\newblock {MMAR}: A challenging benchmark for deep reasoning in speech, audio, music, and their mix.
\newblock \emph{arXiv preprint arXiv:2505.13032}, 2025.
\newblock URL \url{https://arxiv.org/abs/2505.13032}.

\bibitem[Nachmani et~al.(2024)Nachmani, Levkovitch, Hirsch, Salazar, Asawaroengchai, Mariooryad, Rivlin, Skerry-Ryan, and Tadmor~Ramanovich]{nachmani2024spectron}
Eliya Nachmani, Alon Levkovitch, Roy Hirsch, et~al.
\newblock Spoken question answering and speech continuation using spectrogram-powered {LLM}.
\newblock In B.~Kim, Y.~Yue, S.~Chaudhuri, et~al., editors, \emph{International Conference on Learning Representations}, volume 2024, pages 51883--51898, 2024.
\newblock URL \url{https://proceedings.iclr.cc/paper_files/paper/2024/file/e393677793767624f2821cec8bdd02f1-Paper-Conference.pdf}.

\bibitem[Niu et~al.(2025)Niu, Li, Miao, Ge, Zhou, He, Dong, Duan, Ding, Qian, Zhang, Zang, Cao, He, and Wang]{niu2025ovobench}
Junbo Niu, Yifei Li, Ziyang Miao, et~al.
\newblock {OVO-Bench}: How far is your video-{LLMs} from real-world online video understanding?
\newblock In \emph{Proceedings of the IEEE/CVF Conference on Computer Vision and Pattern Recognition}, pages 18902--18913, 2025.
\newblock URL \url{https://openaccess.thecvf.com/content/CVPR2025/html/Niu_OVO-Bench_How_Far_is_Your_Video-LLMs_from_Real-World_Online_Video_CVPR_2025_paper.html}.

\bibitem[{NVIDIA}(2026)]{nvidia2026voicechat}
{NVIDIA}.
\newblock {NVIDIA-NemotronLabs-VoiceChat-11B}.
\newblock Model card, August 2026.
\newblock URL \url{https://huggingface.co/nvidia/NVIDIA-NemotronLabs-VoiceChat-11B}.

\bibitem[{OpenAI}(2024)]{openai2024gpt4o}
{OpenAI}.
\newblock {GPT-4o} system card, 2024.
\newblock URL \url{https://arxiv.org/abs/2410.21276}.

\bibitem[{OpenAI}(2025)]{openai2025gptrealtime}
{OpenAI}.
\newblock Introducing {gpt-realtime} and {Realtime API} updates for production voice agents, August 2025.
\newblock URL \url{https://openai.com/index/introducing-gpt-realtime/}.

\bibitem[{OpenAI}(2026)]{openai2026gptlive}
{OpenAI}.
\newblock Introducing {GPT-Live}, July 2026.
\newblock URL \url{https://openai.com/index/introducing-gpt-live/}.

\bibitem[{OpenBMB}(2025)]{openbmb2025minicpmo26}
{OpenBMB}.
\newblock {MiniCPM-o} 2.6: A {GPT-4o} level {MLLM} for vision, speech and multimodal live streaming on your phone.
\newblock Hugging Face model card, 2025.
\newblock URL \url{https://huggingface.co/openbmb/MiniCPM-o-2_6}.

\bibitem[Quigley(2025)]{quigley2025ultravox}
Keegan Quigley.
\newblock Beyond benchmark-maxxing: Measuring open source models as real-world agents, August 2025.
\newblock URL \url{https://www.ultravox.ai/blog/beyond-benchmark-maxxing-measuring-open-source-models-as-real-world-voice-agents}.

\bibitem[{Qwen Team}(2026)]{qwen2026qwen35}
{Qwen Team}.
\newblock {Qwen3.5}: Towards native multimodal agents, February 2026.
\newblock URL \url{https://qwen.ai/blog?id=qwen3.5}.

\bibitem[Radford et~al.(2023)Radford, Kim, Xu, Brockman, McLeavey, and Sutskever]{Whisper}
Alec Radford, Jong~Wook Kim, Tao Xu, et~al.
\newblock Robust speech recognition via large-scale weak supervision.
\newblock In \emph{International Conference on Machine Learning, {ICML}}, volume 202 of \emph{Proceedings of Machine Learning Research}, pages 28492--28518. {PMLR}, 2023.

\bibitem[Sakshi et~al.(2024)Sakshi, Tyagi, Kumar, Seth, Selvakumar, Nieto, Duraiswami, Ghosh, and Manocha]{sakshi2024mmaumassivemultitaskaudio}
S~Sakshi, Utkarsh Tyagi, Sonal Kumar, et~al.
\newblock {MMAU}: A massive multi-task audio understanding and reasoning benchmark, 2024.
\newblock URL \url{https://arxiv.org/abs/2410.19168}.

\bibitem[Shi et~al.(2026)Shi, Zhou, Lin, Cui, Zeng, Zhou, Wang, Liu, Luo, Wang, and Liu]{ultraevalaudio}
Qundong Shi, Jie Zhou, Biyuan Lin, et~al.
\newblock {UltraEval-Audio}: A unified framework for comprehensive evaluation of audio foundation models.
\newblock \emph{arXiv preprint arXiv:2601.01373}, 2026.
\newblock URL \url{https://arxiv.org/abs/2601.01373}.

\bibitem[Tang et~al.(2024)Tang, Yu, Sun, Chen, Tan, Li, Lu, Ma, and Zhang]{tang2024salmonn}
Changli Tang, Wenyi Yu, Guangzhi Sun, et~al.
\newblock {SALMONN}: Towards generic hearing abilities for large language models.
\newblock In \emph{The Twelfth International Conference on Learning Representations}, 2024.
\newblock URL \url{https://proceedings.iclr.cc/paper_files/paper/2024/hash/476ab8f369e489c04187ba84f68cfa68-Abstract-Conference.html}.

\bibitem[Tang et~al.(2025)Tang, Li, Yang, Zhuang, Sun, Li, Ma, and Zhang]{tang2025videosalmonn2}
Changli Tang, Yixuan Li, Yudong Yang, et~al.
\newblock {video-SALMONN 2}: Caption-enhanced audio-visual large language models.
\newblock \emph{arXiv preprint arXiv:2506.15220}, 2025.
\newblock \doi{10.48550/arXiv.2506.15220}.
\newblock URL \url{https://arxiv.org/abs/2506.15220}.

\bibitem[Tao et~al.(2026)Tao, Zheng, Xu, Du, Shao, Wang, Chen, Jin, Zhu, Yu, Wang, Liu, Qin, Zhang, Yang, and Wang]{tao2026lvomnibench}
Keda Tao, Yuhua Zheng, Jia Xu, et~al.
\newblock {LVOmniBench}: Pioneering long audio-video understanding evaluation for omnimodal {LLMs}.
\newblock \emph{arXiv preprint arXiv:2603.19217}, 2026.
\newblock URL \url{https://arxiv.org/abs/2603.19217}.

\bibitem[{Tongyi Fun Team} et~al.(2025){Tongyi Fun Team}, Chen, Cheng, Deng, Li, Liu, Tan, Wang, Xu, Ye, Zhang, Zhang, and Zhou]{tongyi2025funaudiochat}
{Tongyi Fun Team}, Qian Chen, Luyao Cheng, et~al.
\newblock {Fun-Audio-Chat} technical report.
\newblock \emph{arXiv preprint arXiv:2512.20156}, 2025.
\newblock \doi{10.48550/arXiv.2512.20156}.
\newblock URL \url{https://arxiv.org/abs/2512.20156}.

\bibitem[Tschannen et~al.(2025)Tschannen, Gritsenko, Wang, Naeem, Alabdulmohsin, Parthasarathy, Evans, Beyer, Xia, Mustafa, et~al.]{siglip2}
Michael Tschannen, Alexey Gritsenko, Xiao Wang, et~al.
\newblock Siglip 2: Multilingual vision-language encoders with improved semantic understanding, localization, and dense features.
\newblock \emph{arXiv preprint arXiv:2502.14786}, 2025.

\bibitem[Wang et~al.(2025{\natexlab{a}})Wang, Wu, Li, Yang, Chen, Zhang, and Meng]{wang2025mmsu}
Dingdong Wang, Jincenzi Wu, Junan Li, et~al.
\newblock {MMSU}: A massive multi-task spoken language understanding and reasoning benchmark.
\newblock \emph{arXiv preprint arXiv:2506.04779}, 2025{\natexlab{a}}.
\newblock URL \url{https://arxiv.org/abs/2506.04779}.

\bibitem[Wang et~al.(2025{\natexlab{b}})Wang, Gao, Gu, Pu, Cui, Wei, Liu, Jing, Ye, Shao, Wang, Chen, Zhang, Yang, Wang, Wei, Yin, Li, Cui, Chen, Ding, Tian, Wu, Xie, Li, Yang, Duan, Wang, Hou, Hao, Zhang, Li, Zhao, Duan, Deng, Fu, He, Wang, He, Shi, He, Xiong, Lv, Wu, Shao, Zhang, Deng, Qi, Ge, Guo, Zhang, Zhang, Cao, Lin, Tang, Gao, Huang, Gu, Lyu, Tang, Wang, Lv, Ouyang, Wang, Dou, Zhu, Lu, Lin, Dai, Su, Zhou, Chen, Qiao, Wang, and Luo]{wang2025internvl35}
Weiyun Wang, Zhangwei Gao, Lixin Gu, et~al.
\newblock {InternVL3.5}: Advancing open-source multimodal models in versatility, reasoning, and efficiency.
\newblock \emph{arXiv preprint arXiv:2508.18265}, 2025{\natexlab{b}}.
\newblock \doi{10.48550/arXiv.2508.18265}.
\newblock URL \url{https://arxiv.org/abs/2508.18265}.

\bibitem[Wang et~al.(2025{\natexlab{c}})Wang, Li, Fu, Zhang, Shen, Xie, Li, Sun, and Ma]{wang2025freezeomni}
Xiong Wang, Yangze Li, Chaoyou Fu, et~al.
\newblock {Freeze-Omni}: A smart and low latency speech-to-speech dialogue model with frozen {LLM}.
\newblock In \emph{Proceedings of the 42nd International Conference on Machine Learning}, volume 267 of \emph{Proceedings of Machine Learning Research}, pages 63345--63354. PMLR, 2025{\natexlab{c}}.
\newblock URL \url{https://proceedings.mlr.press/v267/wang25aw.html}.

\bibitem[Wang et~al.(2025{\natexlab{d}})Wang, Liu, Wang, Xu, Wan, Zhang, and Zhao]{wang2025mmduet2}
Yueqian Wang, Songxiang Liu, Disong Wang, et~al.
\newblock {MMDuet2}: Enhancing proactive interaction of video {MLLMs} with multi-turn reinforcement learning.
\newblock \emph{arXiv preprint arXiv:2512.06810}, 2025{\natexlab{d}}.
\newblock \doi{10.48550/arXiv.2512.06810}.
\newblock URL \url{https://arxiv.org/abs/2512.06810}.

\bibitem[Wang et~al.(2025{\natexlab{e}})Wang, Meng, Wang, Zhang, and Zhao]{wang2025proactivevideoqa}
Yueqian Wang, Xiaojun Meng, Yifan Wang, et~al.
\newblock {ProactiveVideoQA}: A comprehensive benchmark evaluating proactive interactions in video large language models.
\newblock \emph{arXiv preprint arXiv:2507.09313}, 2025{\natexlab{e}}.
\newblock URL \url{https://arxiv.org/abs/2507.09313}.

\bibitem[Wu et~al.(2025)Wu, Yan, Hu, Yi, Feng, Tian, Shen, Yu, Zhang, Li, et~al.]{wu2025stepaudio2}
Boyong Wu, Chao Yan, Chen Hu, et~al.
\newblock {Step-Audio} 2 technical report.
\newblock \emph{arXiv preprint arXiv:2507.16632}, 2025.
\newblock \doi{10.48550/arXiv.2507.16632}.
\newblock URL \url{https://arxiv.org/abs/2507.16632}.

\bibitem[Wu et~al.(2026)Wu, Mathews, Cai, Yang, and Wang]{wu2026semantic}
Hang Wu, Sherin~Mary Mathews, Yujun Cai, et~al.
\newblock Semantic-aware adaptive visual memory for streaming video understanding.
\newblock \emph{arXiv preprint arXiv:2605.07897}, 2026.

\bibitem[Wu et~al.(2024)Wu, Li, Chen, and Li]{wu2024longvideobench}
Haoning Wu, Dongxu Li, Bei Chen, et~al.
\newblock {LongVideoBench}: A benchmark for long-context interleaved video-language understanding.
\newblock \emph{Advances in Neural Information Processing Systems}, 37:\penalty0 28828--28857, 2024.

\bibitem[{xAI}(2025)]{xai2025grokvoice}
{xAI}.
\newblock {Grok Voice Agent API}, December 2025.
\newblock URL \url{https://x.ai/news/grok-voice-agent-api}.

\bibitem[{Xiaomi LLM-Core Team}(2025)]{xiaomi2025mimoaudio}
{Xiaomi LLM-Core Team}.
\newblock {MiMo-Audio}: Audio language models are few-shot learners.
\newblock \emph{arXiv preprint arXiv:2512.23808}, 2025.
\newblock \doi{10.48550/arXiv.2512.23808}.
\newblock URL \url{https://arxiv.org/abs/2512.23808}.

\bibitem[Xu et~al.(2025{\natexlab{a}})Xu, Guo, He, Hu, He, Bai, Chen, Wang, Fan, Dang, et~al.]{xu2025qwen2}
Jin Xu, Zhifang Guo, Jinzheng He, et~al.
\newblock {Qwen2.5-Omni} technical report.
\newblock \emph{arXiv preprint arXiv:2503.20215}, 2025{\natexlab{a}}.
\newblock \doi{10.48550/arXiv.2503.20215}.
\newblock URL \url{https://arxiv.org/abs/2503.20215}.

\bibitem[Xu et~al.(2025{\natexlab{b}})Xu, Guo, Hu, Chu, Wang, He, Wang, Shi, He, Zhu, Lv, Wang, Guo, Wang, Ma, Zhang, Zhang, Hao, Guo, Yang, Zhang, Ma, Wei, Bai, Chen, Liu, Wang, Yang, Liu, Ren, Zheng, Men, Zhou, Yu, Yang, Yu, Zhou, and Lin]{xu2025qwen3omni}
Jin Xu, Zhifang Guo, Hangrui Hu, et~al.
\newblock {Qwen3-Omni} technical report.
\newblock \emph{arXiv preprint arXiv:2509.17765}, 2025{\natexlab{b}}.
\newblock \doi{10.48550/arXiv.2509.17765}.
\newblock URL \url{https://arxiv.org/abs/2509.17765}.

\bibitem[Yang et~al.(2025)Yang, Zhang, Hu, Wang, Qian, Wen, Yang, Gao, Dong, and Xu]{yang2025livestar}
Zhenyu Yang, Kairui Zhang, Yuhang Hu, et~al.
\newblock {LiveStar}: Live streaming assistant for real-world online video understanding.
\newblock \emph{arXiv preprint arXiv:2511.05299}, 2025.
\newblock \doi{10.48550/arXiv.2511.05299}.
\newblock URL \url{https://arxiv.org/abs/2511.05299}.

\bibitem[Yao et~al.(2026)Yao, Zhou, Yang, Qin, Hou, Liang, Wang, Cao, Ye, Xie, Gu, Huang, Si, Duan, and Wang]{joyai2026vlinteraction}
Dingyu Yao, Junhao Zhou, Chenxu Yang, et~al.
\newblock {JoyAI-VL-Interaction}: Real-time vision-language interaction intelligence.
\newblock \emph{arXiv preprint arXiv:2606.14777}, 2026.
\newblock \doi{10.48550/arXiv.2606.14777}.
\newblock URL \url{https://arxiv.org/abs/2606.14777}.

\bibitem[Yao et~al.(2023)Yao, Zhao, Yu, Du, Shafran, Narasimhan, and Cao]{yao2023react}
Shunyu Yao, Jeffrey Zhao, Dian Yu, et~al.
\newblock {ReAct}: Synergizing reasoning and acting in language models.
\newblock In \emph{The Eleventh International Conference on Learning Representations}, 2023.
\newblock URL \url{https://arxiv.org/abs/2210.03629}.

\bibitem[Zhang et~al.(2023)Zhang, Li, Zhang, Zhan, Wang, Zhou, and Qiu]{zhang2023speechgpt}
Dong Zhang, Shimin Li, Xin Zhang, et~al.
\newblock {SpeechGPT}: Empowering large language models with intrinsic cross-modal conversational abilities.
\newblock In \emph{Findings of the Association for Computational Linguistics: EMNLP 2023}, pages 15757--15773. Association for Computational Linguistics, 2023.
\newblock \doi{10.18653/v1/2023.findings-emnlp.1055}.
\newblock URL \url{https://aclanthology.org/2023.findings-emnlp.1055/}.

\bibitem[Zhang et~al.(2026)Zhang, Chen, Wu, Li, Zhang, Zhang, Liu, Lin, Peng, Liu, Chng, Yan, Wu, Huang, Yang, and Tian]{zhang2026duplexsla}
Haoyang Zhang, Jun Chen, Donghang Wu, et~al.
\newblock {DuplexSLA}: A full-duplex spoken language model with synchronized speech, language, and action.
\newblock \emph{arXiv preprint arXiv:2605.20755}, 2026.
\newblock \doi{10.48550/arXiv.2605.20755}.
\newblock URL \url{https://arxiv.org/abs/2605.20755}.

\bibitem[Zhao et~al.(2026)Zhao, Yang, Xin, Wang, Rao, Lyu, and Li]{zhao2026omnipro}
Ruixiang Zhao, Jie Yang, Zijie Xin, et~al.
\newblock {OmniPro}: A comprehensive benchmark for omni-proactive streaming video understanding.
\newblock \emph{arXiv preprint arXiv:2605.18577}, 2026.
\newblock URL \url{https://arxiv.org/abs/2605.18577}.

\bibitem[Zhou et~al.(2025)Zhou, Wang, Wu, and Jiang]{zhou2025dailyomni}
Ziwei Zhou, Rui Wang, Zuxuan Wu, et~al.
\newblock {Daily-Omni}: Towards audio-visual reasoning with temporal alignment across modalities.
\newblock \emph{arXiv preprint arXiv:2505.17862}, 2025.
\newblock URL \url{https://arxiv.org/abs/2505.17862}.

\end{thebibliography}
